\documentclass[10pt]{article}
\usepackage[hmargin=2cm,vmargin=2cm]{geometry}      % onecolumn (second format)
\usepackage{graphicx,enumerate,paralist,multicol,url,etoolbox,xcolor}
\usepackage{amsmath,amsfonts,amssymb}
\usepackage[linesnumbered,boxed]{algorithm2e}
\DontPrintSemicolon
\usepackage{newfloat}

\definecolor{armygreen}{rgb}{0.55,0.71,0.0}%{0.29,0.33,0.13}
\definecolor{amber}{rgb}{1.0,0.75,0.0}
\usepackage{authblk}
\title{Multi-Level Evolution Strategies for High-Resolution Black-Box Control}
\author{Ofer M. Shir}
        \affil{Computer Science Department, Tel-Hai College, and Migal Institute, Upper Galilee, Israel }
\author{Xi Xing}
\author{Herschel Rabitz}
		\affil{Department of Chemistry, Princeton University, Princeton NJ 08544, USA}
\date{}
\begin{document}
\SetKwInOut{Input}{input}
\SetKwInOut{Output}{output}

\maketitle

%\SetKwInOut{Input}{input}
%\SetKwInOut{Output}{output}
%\SetFuncSty{texttt}
\begin{abstract}
This paper introduces a multi-level (m-lev) mechanism into Evolution Strategies (ESs) in order to address a class of global optimization problems that could benefit from fine discretization of their decision variables. 
Such problems arise in engineering and scientific applications, which possess a multi-resolution control nature, and thus may be formulated either by means of low-resolution variants (providing coarser approximations with presumably lower accuracy for the general problem) or by high-resolution controls.
A particular scientific application concerns practical Quantum Control (QC) problems, whose targeted optimal controls may be discretized to increasingly higher resolution, which in turn carries the potential to obtain better control yields.
However, state-of-the-art derivative-free optimization heuristics for high-resolution formulations nominally call for an impractically large number of objective function calls. 
Therefore, an effective algorithmic treatment for such problems is needed.
We introduce a framework with an automated scheme to facilitate guided-search over increasingly finer levels of control resolution for the optimization problem, whose on-the-fly learned parameters require careful adaptation.
We instantiate the proposed m-lev self-adaptive ES framework by two specific strategies, namely the classical elitist single-child (1+1)-ES and the non-elitist multi-child derandomized $(\mu_W,\lambda)$-sep-CMA-ES. 
We first show that the approach is suitable by simulation-based optimization of QC systems which were heretofore viewed as too complex to address. We also present a laboratory proof-of-concept for the proposed approach on a basic experimental QC system objective.
\end{abstract}
\textbf{Keywords}: Black-box global optimization, derivative-free search heuristics, multi-resolution, scalability, quantum coherent control, simulation-based optimization, experimental optimization.

\section{Introduction}
\begin{table*}
%\centering Multileveling Nomenclature\\
\begin{small}
\begin{tabular}{l l l}
\hline
Term & Description & Not.\\
\hline
model & single-objective optimization model & $\mathcal{M}$ \\
level index & an indicator for the current problem instance & $\ell$ \\
dimensionality & current scale / resolution; the grid's cardinality & $n_{\ell}$ \\
problem & $n_{\ell}$-dimensional instance of the model $\mathcal{M}$: objective function to be minimized & $\mathcal{P}_{\ell}$ \\
decision space & $n_{\ell}$-dimensional feasible search space, subset of $\mathbb{R}^{n_{\ell}}$ & $\mathcal{X}_{\ell}$ \\
seed & an initial search-point for the ES operating on $\mathcal{P}_{\ell}$ & $\vec{x}^{(0)}_{\ell}$\\
minimizer & a feasible minimizer to the given problem $\mathcal{P}_{\ell}$ & $\vec{x}^{*}_{\ell}$\\
strategy parameters & set of parameters defining ES mutation on $\mathcal{X}_{\ell}$; required for level $(\ell+1)$ & $\mathcal{S}_{\ell}$ \\
leveling-up schedule & the grid enlargement plan as a function of the level index $\ell$ & $\left\{n_{\ell} \right\}$ \\
``fixed-target'' & satisfactory objective function value, serving as termination criterion (tc-1) & $\epsilon$ \\
``stagnation threshold'' & variation threshold of objective function values, serving as termination criterion (tc-2) & $\vartheta$ \\
stagnation sliding-window & number of recent selected individuals to be evaluated per (tc-2) & $w_{\ell}$ \\
initial dimension & the initial level of search; the targeted dimensionality during the first run & $N_i$ \\
final dimension & the ultimate level of search; the targeted dimensionality during the last run & $N_f$ \\
direct dimension & decision-space dimensionality in a direct search (no m-lev) & $N_d$\\
%leveling-up factor & assuming a fixed \textit{schedule},  &  $n_{\ell+1}/n_{\ell}$ \\
null hypothesis & statistical hypothesis: no difference between heuristics in used function calls & $h_0$ \\
\hline
\end{tabular}
\end{small}
\caption{Nomenclature.\label{tab:nomenclature}}
\end{table*}
Control problems are routinely reformulated by higher numbers of variables due to the available advanced state-of-the-art hardware. For instance, modern laser pulse shaping technologies offer experimental apparatus with a growing number of control variables \cite{Weiner00,Hall2016} with the aim of enhancing performance.
At the same time, the challenge of treating global black-box optimization problems subject to an extremely large number of continuous variables translates into algorithmic scalability issues and the inevitable \textit{curse of dimensionality} \cite{Bellman1961}. Unlike optimization of explicit expressions by means of solvers applied to mathematical programs \cite{Boyd}, which are often deployed on convex models comprising thousands of decision variables, \textit{randomized search heuristics} have not been demonstrated to successfully operate on equivalent scales of black-box problems. 
This statement is valid for the broad class of Evolutionary Algorithms (EAs) \cite{Baeck-book}, and the family of Evolution Strategies (ESs) \cite{Baeck2013contemporary,hansen2015,ESchapter2018} -- which are powerful derivative-free heuristics for black-box optimization, and are of particular interest herein. 
Importantly, certain families of real-world problems possess a multi-resolution control nature, i.e., their decision variables may be arranged on a grid. This implicit assumption allows them to be formulated in either coarse or fine scales, meshed in a hierarchy of resolutions. While the fine-scaled problem is generally too complex to tackle, its coarser variants may be solved. 

The broad domain of \textit{multi-resolution} methods \cite{barth2001multiscale} encompasses techniques to automatically solve computational problems by efficient consideration of their underlying scales. 
Multi-resolution problems are common in computational fluid dynamics, nuclear physics and chemistry (i.e., featuring \textit{finite elements}) as well as in digital signal processing (i.e., featuring \textit{wavelets}). The former group is best handled by Multigrid (MG) methods, which were introduced to Applied Mathematics and Computer Science in the form of numerical analysis algorithms to accelerate high-dimensional problem-solving (see, e.g., \cite{hackbusch1985multi,MultigridTutorial}). In essence, MG methods aim to solve a computational problem on the fine scale by moving back and forth between scales, wherein they are primarily applied to solving partial differential equations.
MG methods were later adapted to global optimization targets in order to devise multi-level solvers \cite{MultigridSolvers}. 
Importantly, using the term of multi-level in the MG context is not to be confused with the notion of multi-level in the sense of decomposition \cite{DantzigWolfe}, as sometimes utilized \cite{Tilahun2012}. 
In what follows, we restrict the usage of multi-level to the MG context.
Also, the term Hierarchically Organised ESs has been used \cite{Arnold2006_HOES} in a different context of parameter tuning by means of a bi-level meta-evolution treatment. Despite the terminology resemblance, the latter studies are not relevant to the current work.
%the term Multi-level solvers have been developed with various Computational Intelligence methodologies.
For instance, \textit{multi-level annealing} was presented with the cooling scheme playing the role of identifying promising degrees of freedom at the coarse level \cite{MultigridSolvers}, which translates into introducing new large-scale variables as the temperature is lowered.

Another related research topic is \textit{multi-fidelity optimization}, being primarily concerned with complex systems that may be modelled by high-fidelity functions (the most informative and the best available) and by lowe(er)-fidelity functions that exhibit diminished information at a reduced cost \cite{Multifidelity2012}.
The fidelity is defined either by \textit{hierarchical} or \textit{approximation} models. 
It is decreased accordingly by simplifying the model (``reduced Physics'') and/or coarsening the discretizations within hierarchical models, or by reducing the model's order in approximation models.
The focus in this research thread is mainly on effectively utilizing low-fidelity information to accomplish high-fidelity optimal design, with the primary approaches involving surrogates and meta-modeling, often employing EAs (see, e.g., \cite{EAsMultifidelity2016}).

The concept of extending the array of decision variables during an ongoing optimization process is referred to, within the Evolutionary Computation terminology, as \textit{variable length genotypes}. This concept is rather straightforward in the context of Genetic Programming \cite{Koza1990GP}, where candidate solutions are represented by trees that constantly grow. 
Additionally, this concept was explored by Harvey, who devised a specific Genetic Algorithm dedicated to such a theme (SAGA; see, e.g., \cite{Harvey92speciesadaptation}), with implications mainly in Robotics.
The utilization of other specific MG methods stemming from Computational Intelligence heuristics was reported in multiple studies. 
For example, a multigrid EA was released in \cite{MultigridEAs} to accelerate global optimization of single-objective continuous problems and to boost their accuracy. 
The dimensionality of a given problem remained fixed therein, while the grid/scale of each decision variable ranged from coarse to finer scales, allowing faster convergence and gradual increase of accuracy/precision.
%It followed the original MG concept by addressing a problem in a prescribed dimension of decision variables at various grids/scales, corresponding to degrees of 
Another study \cite{MultigridGAs} proposed a Genetic Algorithm that operates at various dimensions (levels) of an inherently multi-resolution simulation-based optimization problem, where the leveling of the grids is to be manually planned \textit{a priori}. 
A closely-related idea, utilized in a different application and treated by Artificial Immune Systems, was presented in \cite{MultigridAIS}.
Finally, a metamodel-assisted framework for expensive aerodynamics simulation-based optimization problems, either single- or multi-objective, was devised in \cite{Giannakoglou2010}. This comprehensive framework proposed an EA in possibly three multi-level modes -- one of which was multi-level parameterization. The latter relied on available transformations on the decision vectors for switching amongst the levels. This study reported on significant reduction of simulation calls and overall time, yet the independent contribution of the multi-level parameterization was not determined, and the effective roles of the evolutionary operators were not investigated. 
%Another study termed a proposed heuristic as an algorithm for multi-level optimization problems \cite{Tilahun2012}, yet it targeted a different perspective in the context of decomposition of optimization problems \cite{DantzigWolfe}.

It should be stressed that the use of multi-level methods in MG applications is clearly linked to the topic of \textit{evolutionary optimization in dynamic environments} \cite{Branke}, since the search landscape periodically changes, however such consideration of varying dimensionality has not been considered to the best of our knowledge.

The notion of multi-level optimization strategies has not been formally introduced into ESs in the MG perspective, and any related utilization is unknown to us, even when ESs were applied to problems with an extreme number of continuous variables.
The goal of the current study is thus to meet the challenge of introducing multi-level tools into ESs, instantiating it and providing two empirical proofs-of-concept. 
Any proposed multi-level ES would require careful formulation due to the self-adaptive nature of strategy parameters within this family of heuristics.

Importantly, there are certain \textbf{assumptions} on the relevant problems in the current scope of study, considering an objective function, $f~:~\mathbb{R}^d \rightarrow\mathbb{R}$ | %, where $\mathcal{S}$ denotes the feasible region as defined by a given set of constraints:
\begin{compactenum}%[A-1]
%\item The decision variables are defined on a one-dimensional grid, or otherwise may be placed on such.
\item The decision variables, $\vec{x}\in\mathbb{R}^d$, constitute an ordered sequence of elements, $\left\{x_{\jmath} \right\}_{\jmath=1}^{d}$, which is placed on a one-dimensional grid and undergoes some form of interpolation during the evaluation of $f$.
\item $f$ is smooth over the grid and well-defined per each level of the schedule $d \in \left\{ n_{\ell} \right\}$.
\item The model is static in the sense that $f$ does not shift during the course of optimization.
\end{compactenum}
Clearly, discontinuous objective functions, e.g., $$f_{\textrm{dis}}\left(\vec{x} \right)= \left[ \sum_{\imath=2m} x_{\imath}^2 + \sum_{\jmath=2m-1} \left(x_{\jmath}-1\right)^2 \right] \longrightarrow \min,$$ may constitute \textit{deceptive} use-cases for the current framework and are excluded herein. Overall, here are the distilled contributions of the current study:
\begin{enumerate}[i]
\item Formally introducing the notion of m-lev into ESs
\item Devising an upscale operator for ES's strategy parameters
\item Empirically assessing a proposed m-lev approach on a high-dimensional black-box control test-bed, and comparing it to the default approach of addressing the finer-level directly.
\item Deploying the proposed approach on a truly black-box objective function in real-world laboratory settings
\end{enumerate}

In what follows, we present a novel m-lev ES framework, propose the necessary operators, and describe the methodology in detail in Section \ref{sec:MLES}. We then devise instantiations of two specific multi-level ES variants, namely the classical elitist single-child $(1+1)$-ES \cite{Rechenberg,Baeck-book} and the non-elitist multi-child derandomized $(\mu_W,\lambda)$ sep-CMA-ES \cite{HansenDR2PPSN08,Baeck2013contemporary}.
Section \ref{sec:spherePOC} considers the minimization of the unconstrained high-dimensional quadratic model placed on a grid, as a proof-of-concept for the proposed multi-level ES approach.
In what follows, in order to practically assess the proposed framework on meaningful test-cases, we first exercise a simulation-based QC setup -- with the explicit systems under investigation being described in Section \ref{sec:systems}, and the practical observations reported and analyzed in Section \ref{sec:observations}. 
A laboratory-based experimental QC test-case is reported in Section \ref{sec:experiments}.
Finally, we summarize our work and outline directions for future research in Section \ref{sec:conclusions}.
\textit{Nomenclature} is provided in Table \ref{tab:nomenclature} to outline the terminology and notation.

\section{Multi-Level Evolution Strategies}\label{sec:MLES}
We propose an m-lev approach using an ES for high-dimensional problems of a multi-resolution nature.
Consider an optimization model $\mathcal{M}$ formulated on various grid-scales (dimensions) $\left\{ n_{\ell} \right\}$ by means of minimization problems $\left\{ \mathcal{P}_{\ell}: \mathbb{R}^{n_{\ell}} \to \mathbb{R} \right\}$ that are all normalized with a global minimum that has a zero objective function value.
We also assume a self-adaptive ES, operating on $\mathcal{P}_{\ell}$ at dimension $n_{\ell}$, employing a set of strategy parameters $\mathcal{S}_{\ell}$. 
Note that $\mathcal{S}_{\ell}$ may comprise either a scalar (constituting the global-step-size), or a vector (constituting individual step-sizes that represent the distributions' variances). 
Moreover, deploying the ES on $\mathcal{P}_{\ell}$ with a seed point $\vec{x}^{(0)}_{\ell} \in \mathcal{X}_{\ell} \subseteq \mathbb{R}^{n_{\ell}}$, with some performance-based \textit{termination criterion} denoted by $\epsilon$, %aiming to obtain up to a threshold $\epsilon$ with regard to the global minimum (``fixed-target''), 
would result in a randomized heuristic search that outputs a minimizer $\vec{x}^{*}_{\ell} \in \mathcal{X}_{\ell}$ and an adapted strategy $\mathcal{S}_{\ell}$. 
We denote such a self-adapting procedure by \texttt{solveES}.
%wherein the termination criterion is the attainment of an objective function value below the prescribed threshold $\epsilon$.

The main concept behind the proposed multi-level ES is to iteratively increase the dimensionality $n_{\ell}$ upon solving each problem instance $\mathcal{P}_{\ell}$. Each iteration's output, $\left\{ \vec{x}^{*}_{\ell},\mathcal{S}_{\ell} \right\}$, is then lifted-up to the next dimension $n_{\ell+1}$, e.g., by means of a dedicated \textit{upscale operator}, to yield the following iteration's initial components, $\left\{ \vec{x}^{(0)}_{\ell+1},\mathcal{S}_{\ell+1} \right\}$. 
The global step-size, though, is always reduced by a factor of $\sqrt{n_{\ell+1}/n_{\ell}}$, as will be justified below theoretically.
Overall, the following ES operation sets the lifted-up parameters %$\left\{ \vec{x}^{(0)}_{\ell+1},\mathcal{S}_{\ell+1} \right\}$ 
as its seed and as its initial strategy parameter(s), respectively, and accordingly conducts a randomized heuristic $n_{\ell+1}$-dimensional search.

The proposed m-lev approach is summarized as Algorithm \ref{algo:MultilevelES}. 
Throughout this study, we utilize a constant leveling-up \textit{schedule} for simplicity, and fix it to $\forall \ell ~ n_{\ell+1}/n_{\ell}=2$ (see discussion in Section \ref{sec:conclusions}).
The termination criterion per level is assumed to be implemented within \texttt{solveES} and is thus left abstract in Algorithm \ref{algo:MultilevelES}; it will be discussed per our implementations in Section \ref{sec:lcriteria}.
%\begin{figure}
%\centering
%\includegraphics[width=0.7\columnwidth]{ML_algo_main.eps}
%\caption{M-lev Evolution Strategy operating on a constant schedule with $s(\ell) = n_{\ell+1}/n_{\ell}=2$ and some performance-based leveling-up criterion (denoted by $\epsilon$ but unspecified herein).\label{algo:MultilevelES}}
%\end{figure}
\IncMargin{1em}
\begin{algorithm}
%\caption{Multilevel Evolution Strategy \textit{featuring} a fixed schedule with factor $n_{\ell+1}/n_{\ell}=2$. \label{algo:MultilevelES}}
\caption{An m-lev Evolution Strategy operating on a given \textit{schedule} $\left\{n_{\ell} \right\}$ and some \textit{termination criterion} (denoted by $tc$).\label{algo:MultilevelES}}
%\Input{$\textrm{initial_dim} n_i,~ \textrm{final_dim} n_f$}
\Input{$ \textrm{problemModel} ~\mathcal{M},~ \textrm{schedule} ~\left\{n_{\ell} \right\},~ \textrm{finalDim} ~N_f,~ \textrm{tc\_object} ~tc $ }
\Output{ minimizer $\vec{x}^{*} \in \mathbb{R}^{N_f}$}
$\ell \leftarrow 1$\;
%$n_{\ell} \leftarrow N_i$\;
$\vec{x}^{(0)}_{\ell} \leftarrow$\texttt{randomInit}$\left(\mathcal{M},n_{\ell} \right)$ \; %/* Guess initial solution */\;
$\mathcal{S}_{\ell} \leftarrow$\texttt{initStrategy}$\left(\mathcal{M},n_{\ell} \right)$ \; %/* Initialize strategy parameters */\;
\While{$ n_{\ell} \leq N_f $} {
$\mathcal{P}_{\ell}\longleftarrow$\texttt{formProblem}$\left(\mathcal{M},n_{\ell} \right)$ \; %~~~ /* formulate an instance */ \;
\If{ $\ell>1$ } {
$\vec{x}^{(0)}_{\ell} \leftarrow$\texttt{upscale}$\left(\vec{x}^{*}_{\ell-1},n_{\ell} \right)$ \;
$\mathcal{S}_{\ell} \leftarrow$\texttt{upscale}$ \left(\mathcal{S}_{\ell-1},n_{\ell} \right)$\;
$\sigma_{\ell} \leftarrow\frac{\sigma_{\ell-1}}{\sqrt{n_{\ell}/n_{\ell-1}}}$ \;
} 
$\left\{ \vec{x}^{*}_{\ell},\mathcal{S}_{\ell} \right\} \longleftarrow$\texttt{solveES}$\left(\mathcal{S}_{\ell},\mathcal{P}_{\ell},\vec{x}^{(0)}_{\ell},tc \right)$ \; %~~~ /* solve up to threshold $\epsilon$ */ \;
% $n \leftarrow \left(2n < n_f \right)~ ?~ 2n~ :~ n_f $\;
\lIf{ $n_{\ell} == N_f$ } { \Return { $\vec{x}^{*}_{\ell}$ } }
%\lElseIf{ $2n_{\ell} \leq N_f$ } { $n_{\ell+1} \leftarrow 2n_{\ell}$}
%\lElse{ $n_{\ell+1} \leftarrow N_f $}
\lElse{ $\ell \leftarrow \ell + 1$ } %\;
 }
\Return{ $\vec{x}^{*}_{\ell-1}$ }
%\BlankLine
\end{algorithm}
\DecMargin{1em}
The procedure entitled \texttt{formProblem} refers to the explicit formulation of the given optimization problem $\mathcal{M}$ on a grid of size $n_{\ell}$. 
For typical MG problems, this formulation is straightforward and mainly involves numerical adjustment to the grid size and its boundaries.
\subsection{Leveling-up Criteria}\label{sec:lcriteria}
Given the black-box nature of the search landscape, assumptions on the objective function values are needed, to some extent, in order to devise criteria for automated leveling-up.
We consider the following three straightforward \textit{termination criteria}:
\begin{enumerate}[(tc-1)]
\setcounter{enumi}{-1}
\item A fixed budget is allocated per each level.
\item The attainment of an objective function value below a prescribed threshold denoted as $\epsilon$ (also known as ``fixed-target'' optimization).
\item The stagnation of the search as indicated by some \textit{statistics} of the objective function values recorded in a predefined \textit{sliding window}. Particularly, we consider the search to be stagnated when the \textit{range} of the values within the window are below a threshold denoted by $\vartheta$. The sliding window's size is denoted by $w_{\ell}$.
\end{enumerate}
The usage of (tc-0) is meant for zero-assumption black-box optimization problems with possibly expensive objective function calls and unknown noise distributions.
The policy in determining the budget allocation per level is likely to stem from operational considerations (e.g., costs, duration, etc.), and could be analyzed in light of the \textit{multi-armed bandit problem} \cite{powell2013optimal}. 
Next, (tc-1) is applicable when the scale of the objective function values is known \textit{a priori}. It is often encountered in black-box optimization, especially when the objective function has a known lower bound, or is \textit{normalizable} as a whole.
At the same time, (tc-2) is a commonly utilized criterion with practical efficacy in heuristics. However, the effectiveness of (tc-2) is dependent upon the sensitivity of the search landscape's attraction basins. Unknown sensitivity is likely to render the statistical operation over the sliding window useless. Importantly, the applicability of (tc-2) is prone to be limited in real-world settings, mainly due to the existence of noise, which renders profiling attempts problematic. 
\subsection{Leveling-Up the Parameters}
We elaborate here on the update scheme per each leveling-up step:
\paragraph*{The Upscale Operator}~A straightforward treatment for the required task of upscaling the decision variables' vectors is to conduct standard \textit{interpolation} \cite{NumericalRecipes}, while fixing the variables at the edges.
We consider the following variants, listed in ascending order of time and space complexity:
\begin{compactenum}[(U-1)]
\item Nearest neighbor: the simplest form of interpolation, setting the value of the nearest sample grid point
\item Linear: setting linear interpolants between each pair of grid points
\item Cubic: setting shape-preserving piecewise cubic interpolants based on the neighboring grid points (at least 4 grid points are needed)
%\item Spline: 
\end{compactenum}
Boundary conditions enforcement may be implemented by any of the conventional schemes; a particular case will be reported in one of the illustrated systems. 
\paragraph*{Global Step-Size Level-Update}~Theoretical results devising the optimal step-size for an elitist single-child ES operating with the so-called \textit{1/5th success-rule} and targeting an unconstrained $n_{\ell}$-dimensional quadratic model (also known as the Sphere function), are available following the work of Rechenberg \cite{Rechenberg,Beyer}:
\begin{equation}
\label{eq:1p1_optimalstep}
\displaystyle \sigma^{*}_{\ell}\left(\vec{x} \right) \approx 1.224 \cdot \frac{R}{n_{\ell}},
\end{equation}
where $R=\sqrt{f_{\textrm{Sphere}}\left(\vec{x} \right)}$ (see Eq.\ \ref{eq:sphere}).
In the current MG perspective, assuming that the decision variables are simply duplicated per each leveling-up step between $n_{\ell}$ to $n_{\ell+1}$,  the quadratic modeling is subject to increasing the \textit{objective function value} by a factor of ${n_{\ell+1}/n_{\ell}}$. 
\textbf{Since the optimal step-size is proportional to $R/n_{\ell}$, the step-size should be reduced by a factor of $\sqrt{n_{\ell+1}/n_{\ell}}$ in each leveling-up}.
Keeping in mind that the validity of the 1/5th rule is broad, and extends to nonspherical fitness landscapes that may be locally approximated by a \textit{substitute sphere} \cite{Beyer}, we follow this theoretical argumentation to formulate our leveling-up step-size update scheme as the reduction by a factor of $\sqrt{n_{\ell+1}/n_{\ell}}$.

\subsection{Instantiations}
We instantiate the proposed multi-level ES approach by means of two strategies, both holding a single search-point and a single set of strategy parameters in each iteration -- a choice which allows for straightforward multi-level implementation.

\medskip

\paragraph*{\textbf{m-lev-$(1+1)$-ES}}~(denoted \textbf{m-lev-1p1}) The elitist single-child $(1+1)$-ES \cite{Rechenberg,Baeck-book} has become a benchmark strategy due its simplicity, fine performance, but above all, due to the broad understanding of the theory behind its operation \cite{Beyer}. 
This self-adaptive hill-climber adheres to the class of \textit{scalar-driven mutations} within ESs, since its mutation operates only with a global step-size $\sigma_{\ell}$, $$\vec{x}^{\prime}_{\ell} = \vec{x}_{\ell} + \sigma_{\ell} \cdot \mathcal{N} \left(\vec{0},\mathbf{I}\right),$$
and therefore holds $\mathcal{O}(1)$ strategy parameters: $$\mathcal{S}^{(1+1)\textrm{-ES}}_{\ell} = \left\{\sigma_{\ell} \right\}.$$
We consider the default heuristic \cite{Baeck-book} operating in a search-space of dimension $n_{\ell}$, which updates the global step-size according to the 1/5th success-rule: 
Adjustment of $\sigma_{\ell}$ is performed every $n_{\ell}$ mutations, whereas the success-rate is measured over the past $10n_{\ell}$ mutations, and the update constant is set to $c=0.817$ \cite{Baeck-book}.

Importantly, in the m-lev context, the \textit{upscale operator} is applied to the vector of decision variables.
The set of strategy parameters, $\mathcal{S}^{(1+1)\textrm{-ES}}_{\ell}$, comprises the global step-size alone, which is reduced by a factor of $\sqrt{n_{\ell}/n_{\ell-1}}$ each leveling-up. 

\medskip

\paragraph*{\textbf{m-lev-$(\mu_W,\lambda)$-sep-CMA-ES}}~(denoted \textbf{m-lev-sepC}) The sep-CMA-ES \cite{HansenDR2PPSN08} is a modern, derandomized strategy, which was devised to reduce the complexity of the renowned $(\mu_W,\lambda)$-CMA-ES into linear space and time orders in $n_{\ell}$. Its mutation operates with a global step-size, $\sigma_{\ell}$, as well as a vector of individual step-sizes, $\vec{d}_{\ell} \in \mathbb{R}^{n_{\ell}}$, upon utilizing a diagonalized matrix $\mathbf{D}_{\ell}=\textrm{diag} \left(\vec{d}_{\ell} \right)$:
$$\vec{x}^{\prime}_{\ell} = \vec{x}_{\ell} + \sigma_{\ell} \cdot \mathbf{D}_{\ell} \mathcal{N} \left(\vec{0},\mathbf{I}\right).$$
Also, in order to facilitate accumulation of past search information, two auxiliary vectors (entitled \textit{evolution paths} and denoted $\vec{p}_s$, $\vec{p}_c$) are iteratively updated and utilized in the adaptation of $\sigma_{\ell}$ and $\vec{d}_{\ell}$.
This heuristic thus adheres to the class of \textit{vector-driven mutations} within ESs, since it holds $\mathcal{O}\left( n_{\ell}\right)$ strategy parameters. 
In our m-lev method, we choose to \textbf{reset the evolution paths} $\left\{\vec{p}_s,\vec{p}_c \right\}$ each leveling-up, and therefore to define the inherited multi-level strategy parameters as follows: 
$$\mathcal{S}^{(\mu_W,\lambda)\textrm{-sepC}}_{\ell} = \left\{\sigma_{\ell},~\vec{d}_{\ell} \right\}.$$
The strategy parameters are handled as follows -- the \textit{upscale operator} is applied to the vector of individual step-sizes, $\vec{d}_{\ell}$, whereas the global step-size is reduced by a factor of $\sqrt{n_{\ell}/n_{\ell-1}}$ each leveling-up.
In parallel, the \textit{upscale operator} is applied to the vector of decision variables.
For population sizing, our framework follows the recommended default values:
\begin{equation}\label{eq:popsize}
\displaystyle \mu_{\ell}=\lfloor \lambda_{\ell}/2 \rfloor,~~~\lambda_{\ell}=4+\lfloor 3\cdot \ln(n_{\ell}) \rfloor.
\end{equation}
Also, all other parameters are set to the recommended values \cite{HansenDR2PPSN08} per each level $\ell$.

\section{Preliminary: Quadratic Model Placed on a Grid}\label{sec:spherePOC}
Artificially placing the unconstrained quadratic model (also known as the Sphere function) on a high-dimensional grid constitutes a \textit{trivial multi-resolution use-case} due to its isotropic nature and lack of local traps, and yet, targeting it in a \textbf{black-box perspective} serves as a synthetic proof-of-concept for exploring the proposed multi-level ES instantiations.
In other words, the current use-case is an \textit{exercise}, rather than a test-case, and it is reported herein mainly because of the availability of ESs' theoretical results concerning it.

Formally, the $n_{\ell}$-dimensional Sphere is defined by means of the following objective function, subject to minimization:
\begin{equation}
\label{eq:sphere}
\displaystyle f_{\textrm{Sphere}}\left( \vec{x} \right) = \sum_{\imath=1}^{n_{\ell}} x_{\imath}^2 \longrightarrow \min.
\end{equation}
Notably, the current formulation necessarily dictates the deterioration of the objective function values per each leveling-up. 
ESs' operation on the Sphere function is well-understood, and certain results may provide a valuable reference for the behavior of the proposed m-lev-1p1. In particular, we adhere to Eq.~(\ref{eq:1p1_optimalstep}) for obtaining the theoretically-optimal global step-size values.
\subsection{Simulations Planning}
Here, we consider minimizing the Sphere function at a final dimensionality of $N_f=10^4$, by applying the direct $(1+1)$-ES and $(\mu,\lambda)$-sep-CMA-ES, and comparing their performance to the derived multi-level variants, m-lev-1p1 and m-lev-sepC, respectively. The initial dimensionality is set to $N_i=10$. 

In practice, initial candidate solutions were randomly and uniformly generated within $\left[-5,+5 \right]^{N_i}$. The initial global step-size was set to $\sigma_0=\frac{10}{3}$. All three upscale operators, (U-1)-(U-3), were tested for the multi-level ES variants.
The m-lev-sepC starts with a $(5_W,10)$ strategy on $N_i$ and concludes with a $(15_W,31)$ strategy for the full-scale problem on $N_f$. %(which is the population size for the direct strategy applied to the full-scale problem).

Regarding termination criteria, we utilize both (tc-1) and (tc-2): the ``fixed-target'' was set to $\epsilon=0.05$ per (tc-1), and the stagnation threshold was set to $\vartheta=10^{-4}$ per (tc-2). 
The sliding-window's size was set to $w^{(1+1)\textrm{-ES}}_{\ell}= \lceil 100 \cdot \log_{10}(n_{\ell})\rceil$ for the m-lev-1p1, and to $w^{(\mu_W,\lambda)\textrm{-sepC}}_{\ell}=\lceil \lambda_{\ell} \cdot \log_{10}(n_{\ell})\rceil$ for the m-lev-sepC.

Reported comparisons account for the totally utilized number of function evaluations to reach the fixed-target per (tc-1), or the number of evaluations to reach stagnation per (tc-2) as long as the objective function value is below the fixed-target.

\subsection{Practical Observation per (tc-1)}
\begin{figure*}
\centering
\includegraphics[width=0.5\columnwidth]{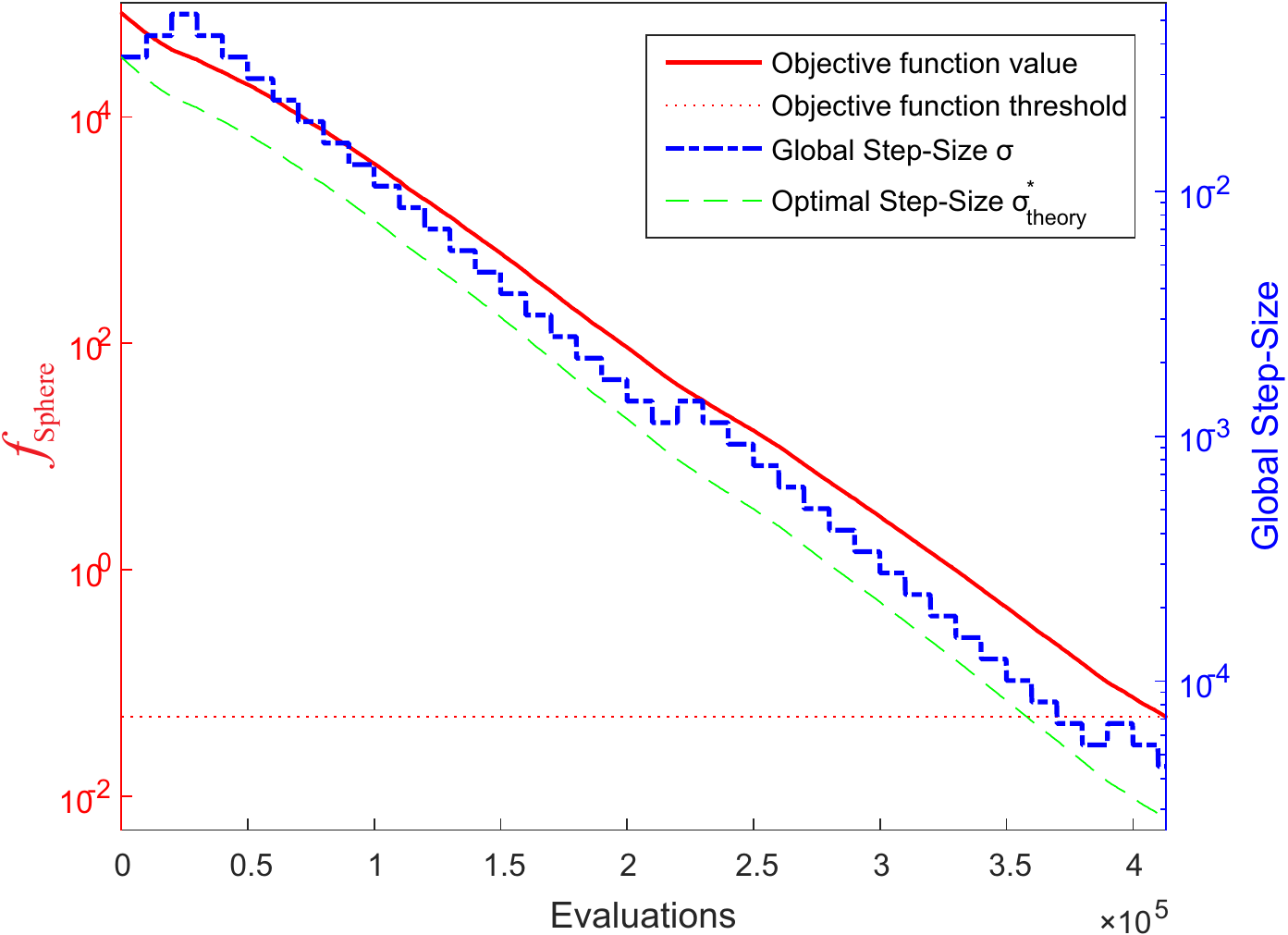}
\includegraphics[width=0.48\columnwidth]{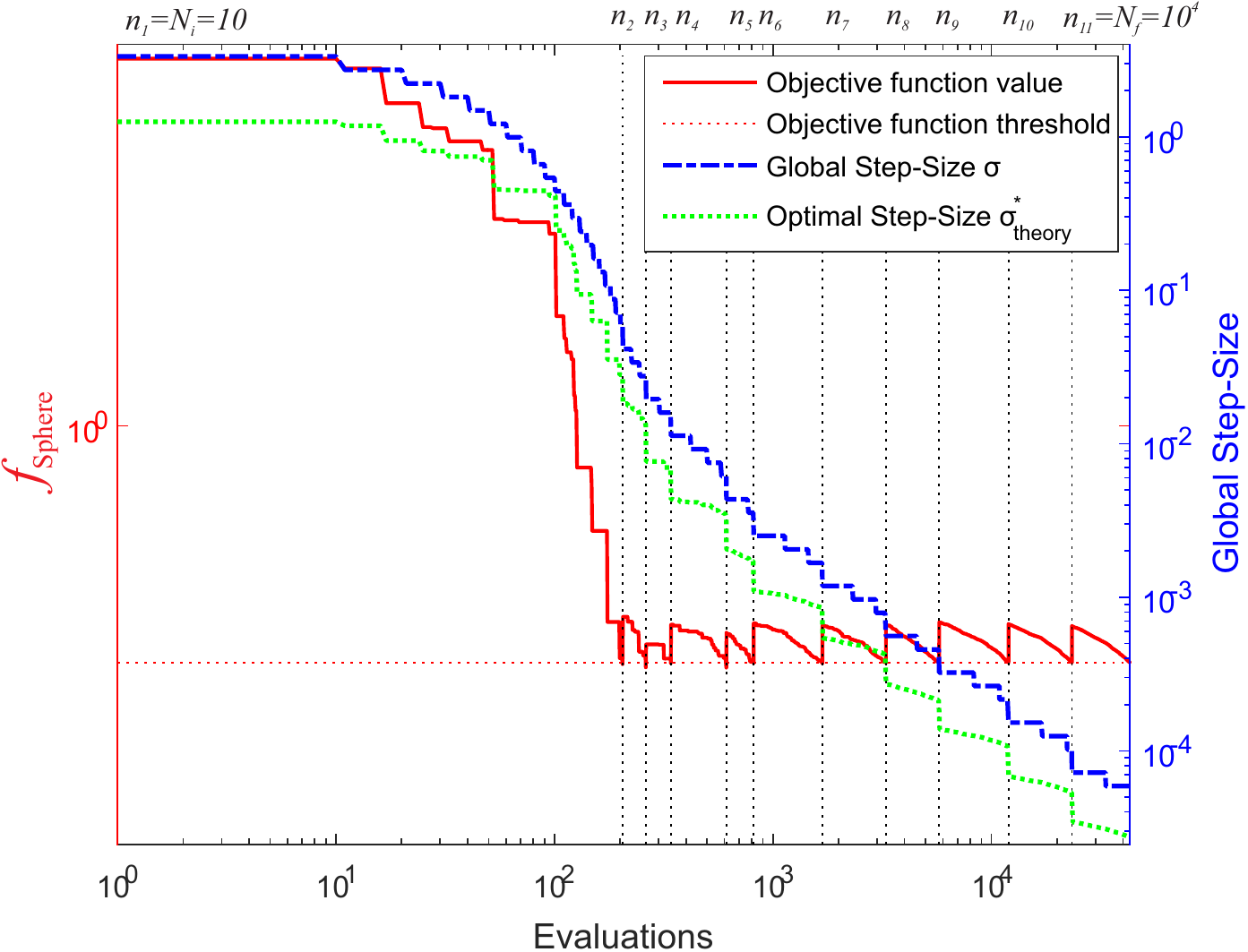}
 \caption{Median runs of $(1+1)$-ES and m-lev-1p1 applied to $f_{\textrm{Sphere}}$ with $N_f=10^4$, \textbf{using (tc-1)} at a fixed-target of $\epsilon=0.05$. [LEFT, logarithmic scale for the y-axis] the \textit{direct variant} on the full-scale problem, versus [RIGHT, log-log scale] the multi-level variant employing (U-3), starting at $N_i=10$, with vertical dashed lines that represent each leveling, corresponding to the displayed $n_{\ell}$ values (top x-axis).
The objective function values (left y-axis) and global step-size (right y-axis) are depicted as a function of the utilized evaluations. 
The theoretically-optimal global step-size $\sigma^{*}$ is calculated using Eq.\ (\ref{eq:1p1_optimalstep}). \label{fig:1p1_sphere}}
\end{figure*}
\begin{figure*}
\centering
\includegraphics[width=0.5\columnwidth]{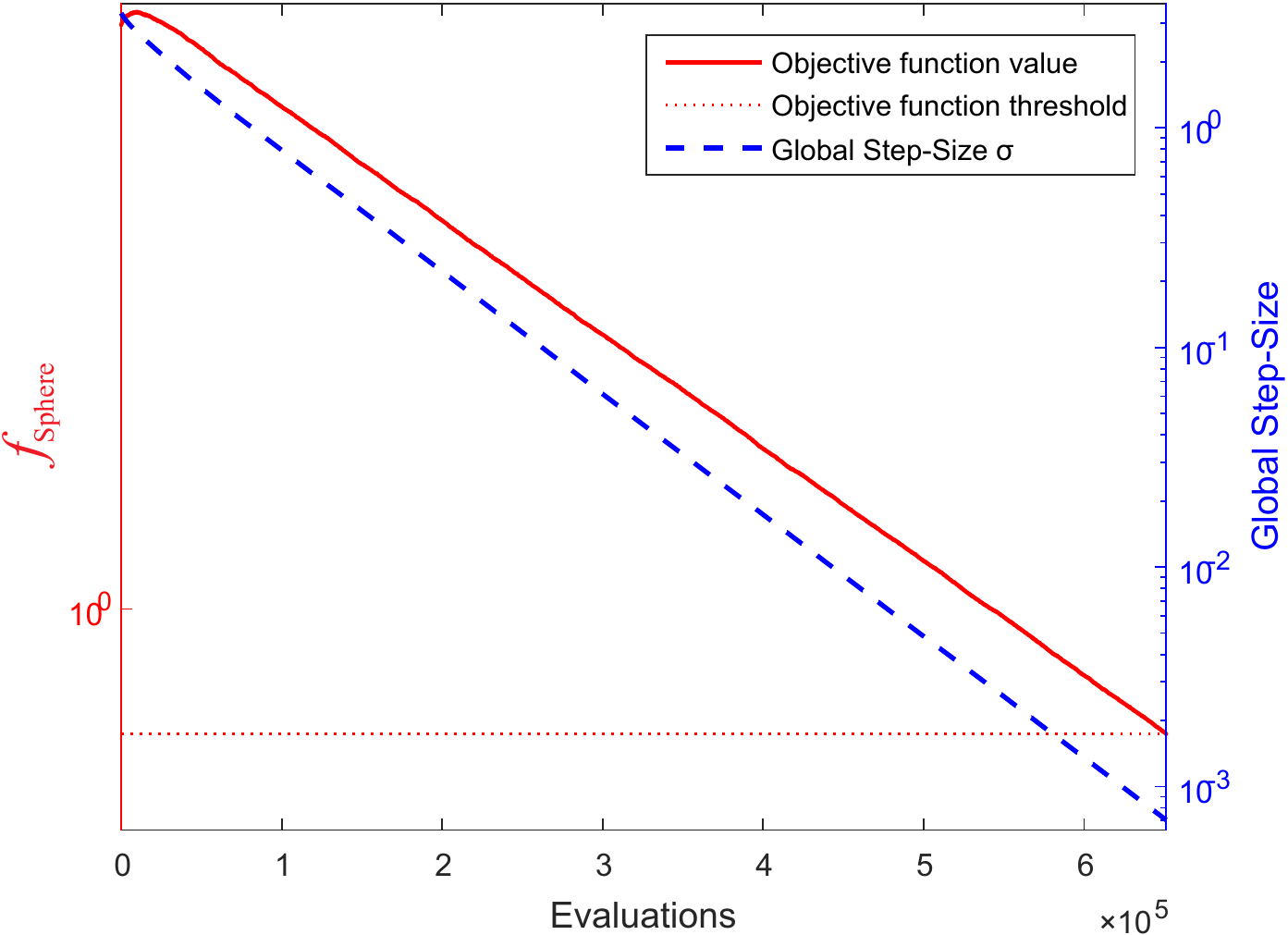}
\includegraphics[width=0.48\columnwidth]{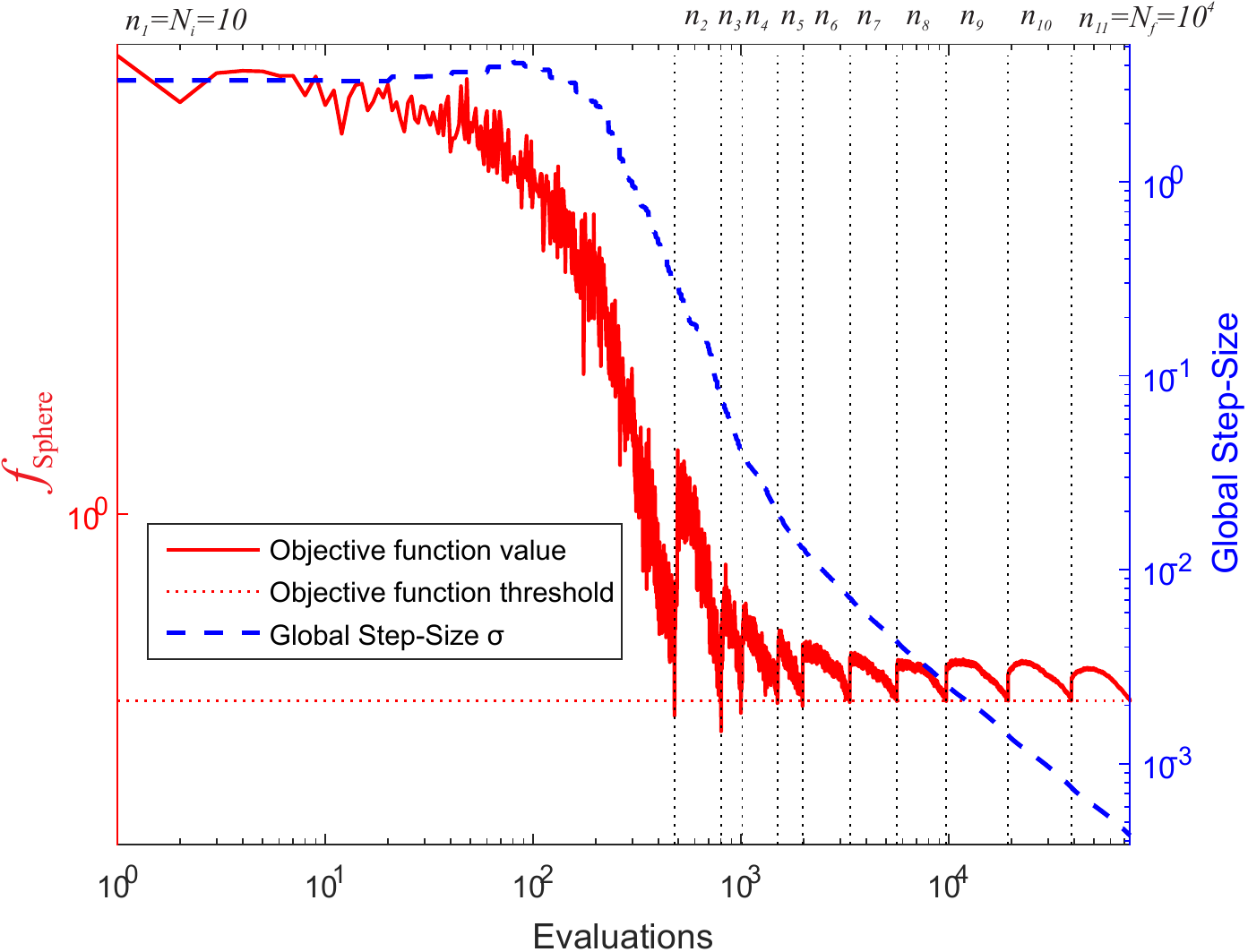}
 \caption{Median runs of $(\mu_W,\lambda)$-sep-CMA-ES and m-lev-sepC applied to $f_{\textrm{Sphere}}$ with $N_f=10^4$, \textbf{using (tc-1)} at a fixed-target of $\epsilon=0.05$. [LEFT, logarithmic scale for the y-axis] the \textit{direct variant} on the full-scale problem, versus [RIGHT, log-log scale] the multi-level variant employing (U-2), starting at $N_i=10$, with vertical dashed lines that represent each leveling, corresponding to the displayed $n_{\ell}$ values (top x-axis).
The objective function values (left y-axis) and global step-size (right y-axis) are depicted as a function of the utilized objective function evaluations. \label{fig:r2d2_sphere}}
\end{figure*}
Fig.~\ref{fig:1p1_sphere} presents the median runs (i.e., 50\%-tile considering the number of objective function evaluations) of the \textit{direct} $(1+1)$-ES versus the m-lev-1p1, depicting the objective function and global step-size values as a function of evaluations. Additionally, the theoretically-optimal global step-size $\sigma^{*}$ is calculated using Eq.~(\ref{eq:1p1_optimalstep}) and depicted over those plots.
Fig.~\ref{fig:r2d2_sphere} presents median runs of the \textit{direct} $(\mu_W,\lambda)$-sep-CMA-ES versus the m-lev-sepC, depicting the objective function and global step-size values as a function of evaluations.
In both figures, vertical dashed lines are displayed for the multi-level runs to represent each leveling, corresponding to the displayed $n_{\ell}$ values. Also, log-log scales are chosen for those plots to accommodate fine observation of the leveling.
Finally, Fig.~\ref{fig:boxplot_sphere} provides statistical boxplots reflecting 30 runs of each m-lev variant per either (tc-1) or (tc-2), employing all three \textit{upscale} operators, alongside the \textit{direct} ESs.

Next, we discuss the \textit{statistics} of the recorded runs, considering the utilized number of objective function calls to reach $\epsilon$.
Average values are accompanied with t-distribution confidence intervals at the 99\% level.
The \textit{direct} variants required on average at least $4\cdot 10^5$ function evaluations for minimizing $f_{\textrm{Sphere}}$: 
$(1+1)$-ES utilized $\approx 4\cdot 10^5 \pm 3,000$ calls, and the $(\mu_W,\lambda)$-sep-CMA-ES utilized $\approx 6.5\cdot 10^5 \pm 1,500$ calls. 
The multi-level variants required significantly fewer calls:
m-lev-1p1 variants required on average $\approx 4\cdot 10^4 \pm 1,000$ calls, while the m-lev-sepC required $\approx 3\cdot 10^4 \pm 300$ calls. 
%$3 \sim 4 \cdot 10^4$ across the different variants (with moderate standard deviations of $\lesssim 1000$). 
For all cases, median values were observed in practice to hit the reported mean values.
A first conclusion states that multileveling of the $f_{\textrm{Sphere}}$ proved successful and obtained a speed-up of one order of magnitude in evaluations.

A second conclusion may be drawn with regard to the global step-size of the elitist single-child m-lev-1p1. 
Evidently, the strategy's global step-size systematically follows the pattern of the optimal step-size while keeping a steady small gap: the observed gap median for (U-1) was $\approx 3.5\cdot 10^{-5}$, whereas the gap median for (U-2),(U-3) was $\approx 6\cdot 10^{-5}$.
Importantly, both the gap and the pattern are consistently maintained throughout the multi-level scheme.

Upon further examining Fig.~\ref{fig:boxplot_sphere} \textbf{in light of (tc-1)}, it is evident that m-lev-sepC operated best on $f_{\textrm{Sphere}}$ when utilizing (U-2), unlike m-lev-1p1 which operated similarly with the three operators. 
Additionally, statistical comparisons were drawn from the numerical results of the three upscale operators per each ES variant. 
Following a Friedman test to ensure that at least one variant has significant differences with respect to the other, statistical Mann-Whitney U-tests (Wilcoxon's rank sum tests) were conducted across all approaches with a \textit{null hypothesis} $h_0$ stating that there was no performance difference between them in terms of the averages of the necessary objective function calls.
Those comparisons concluded that $h_0$ was always rejected at the 5\% significance level for the m-lev-sepC, and thus, the good performance statement concerning (U-2) is statistically valid. At the same time, $h_0$ was never rejected for the m-lev-1p1, supporting the observation of similar performance. 
Both statements constitute a third conclusion for this proof-of-concept when using (tc-1).

\subsection{Practical Observation per (tc-2)}
\begin{figure*}
\centering
\includegraphics[width=0.99\columnwidth]{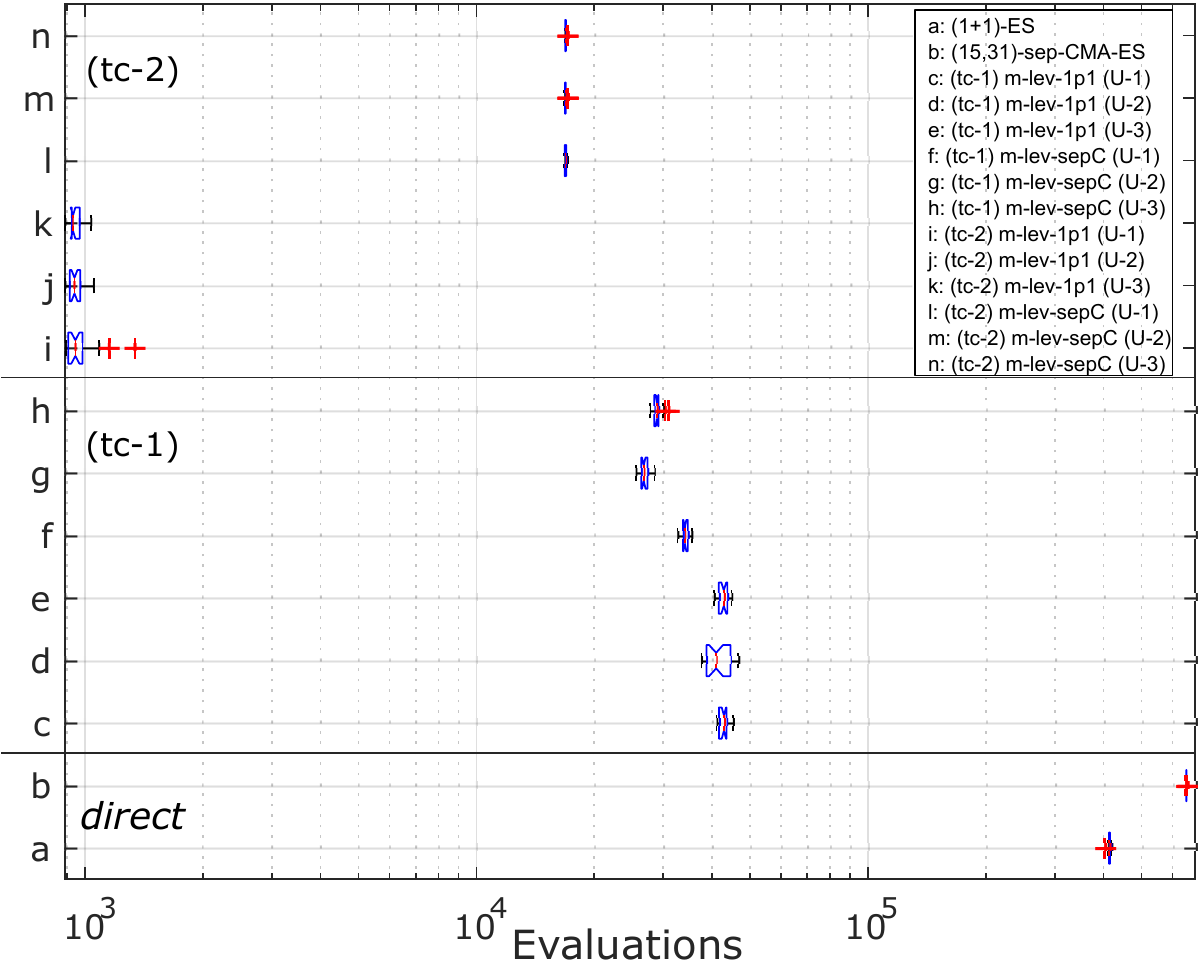}
 \caption{Statistical boxplots accounting for 30 runs of minimizing the Sphere function by all 14 ESs instantiations: \textit{direct} variants, and m-lev variants with all three upscale operators, (U-1)-(U-3), either using (tc-1) at a fixed-target of $\epsilon=0.05$ , or using (tc-2) with a stagnation threshold of $\vartheta=10^{-4}$. Each m-lev starts at $N_i=10$ and levels-up to $N_f=10^4$. \label{fig:boxplot_sphere}}
\end{figure*}

The utilization of (tc-2) proved successful, and the speed-up in evaluations was kept when comparing to the \textit{direct} ES.
Fig.~\ref{fig:boxplot_sphere} encompasses the explicit statistical boxplots, depicting both (tc-1) and (tc-2) alongside the direct variants.
Clearly, the usage of (tc-2) required even fewer evaluations when compared to (tc-1):
m-lev-1p1 variants required on average $\approx 3,175 \pm 30$ calls, while the m-lev-sepC required $\approx 1.725\cdot 10^4 \pm 50$ calls | both exhibiting tight distributions.
Further statistical tests indicated that the performance of m-lev-1p1 was significantly better than m-lev-sepC (that is, it reached stagnation, within the global optimum, significantly faster). However, the performance of the three upscale operators per each m-lev instantiation was observed to be similar (that is, $h_0$ was never rejected among the three upscale operators).
Regarding m-lev-1p1, and equivalently to (tc-1), its global step-size consistently followed the pattern of the theoretically optimal step-size $\sigma^{*}$, as previously observed and reported.

\begin{figure*}
\centering
\includegraphics[width=0.48\columnwidth]{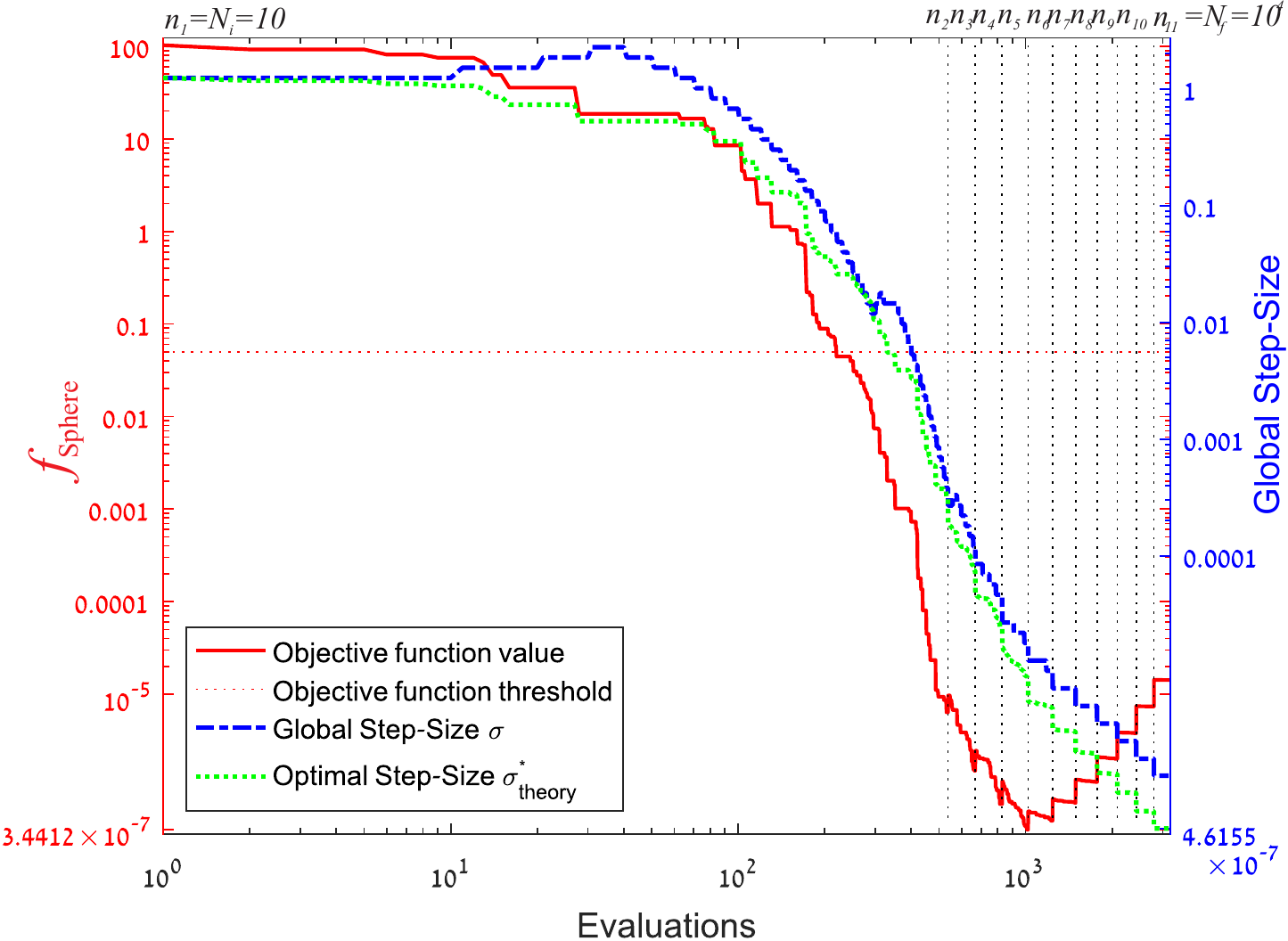}
\includegraphics[width=0.48\columnwidth]{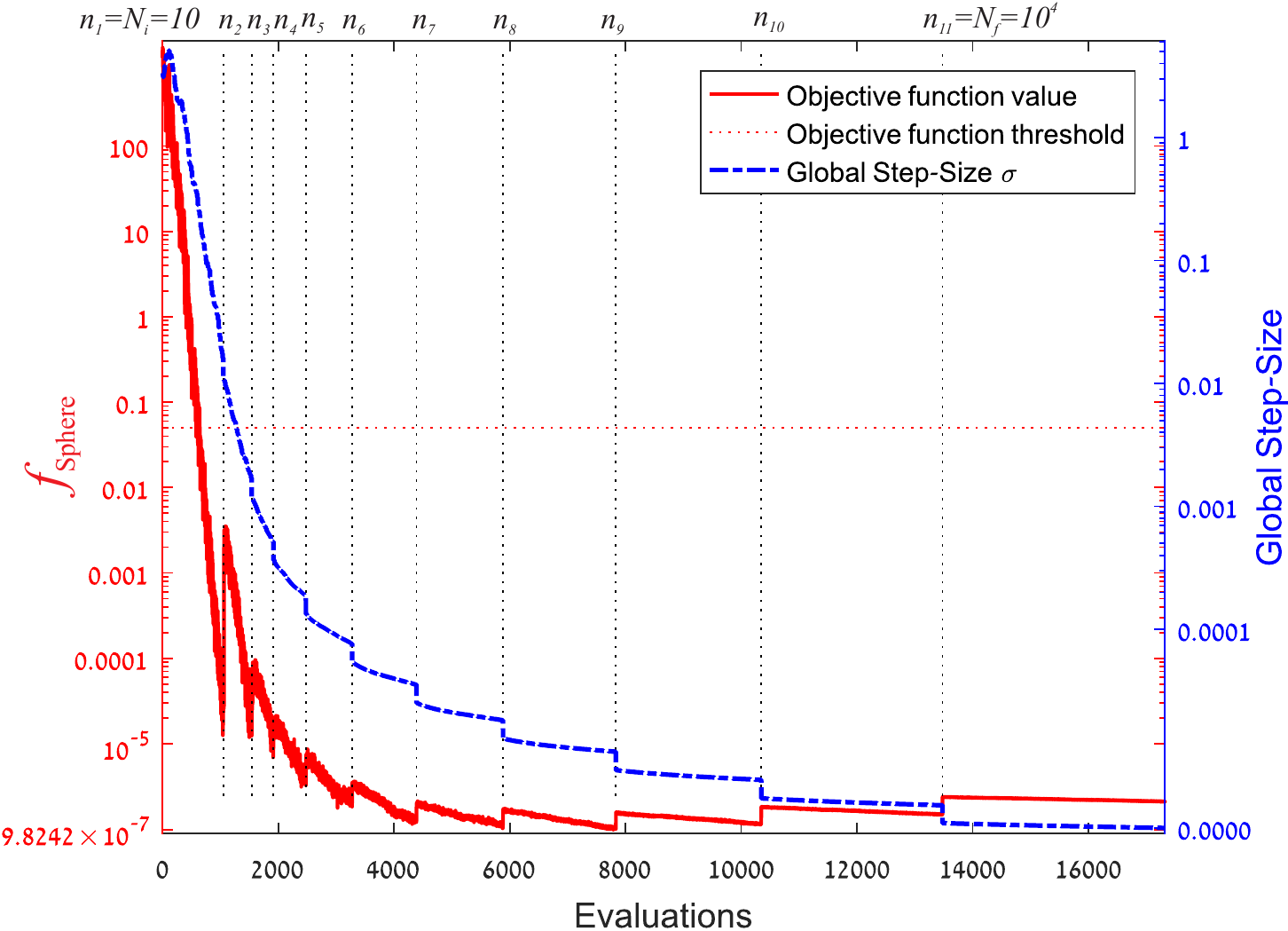}
 \caption{Median runs of m-lev-1p1 and m-lev-sepC applied to $f_{\textrm{Sphere}}$ with $N_f=10^4$, \textbf{using (tc-2)} at a stagnation threshold of $\vartheta=10^{-4}$, starting at $N_i=10$. Vertical dashed lines represent each leveling, corresponding to the displayed $n_{\ell}$ values (top x-axis).
The objective function values (left y-axis) and global step-size (right y-axis) are depicted as a function of the utilized evaluations. 
[LEFT, log-log scale] m-lev-1p1 employing (U-3), whereas the theoretically-optimal global step-size $\sigma^{*}$ is calculated using Eq.\ (\ref{eq:1p1_optimalstep}). 
[RIGHT, log-log scale] m-lev-sepC employing (U-2).
The objective function threshold per (tc-1) is depicted as a reference using a dashed horizontal line.\label{fig:sphere_THF}}
\end{figure*}

Nevertheless, the \textbf{qualitative profile of the runs per (tc-2) was evidently different than (tc-1)}. 
Fig.~\ref{fig:sphere_THF} presents median runs, per selected upscale operators, of m-lev-1p1 alongside m-lev-sepC. 
The objective function threshold per (tc-1), $\epsilon=0.05$, is depicted in Fig.~\ref{fig:sphere_THF} as a reference to reflect the quality of the attained solutions when using the current termination criterion.
Notably, both m-lev variants suffer deteriorations in objective function values in the late stages, likely due to the combined effect of further leveling-up with a diminished step-size. 
This behavior is consistent throughout all runs. It is incomparable to (tc-1), since this effect occurs deeper in the attraction basic of the global optimum, far beyond the termination of (tc-1)-driven runs. 
It is explained by the absence of stagnation in this unconstrained convex landscape until the search deeply penetrates the attraction basin.

Lastly but importantly, the deployment of (tc-2) required an effort in setting its defining parameters, $\left\{w,\vartheta\right\}$, to work, which in turn, exhibit a strong impact on the heuristic behavior. 
The reported configurations for $w^{(1+1)\textrm{-ES}}_{\ell}$ and $w^{(\mu_W,\lambda)\textrm{-sepC}}_{\ell}$ are stable problem-dependent settings that may not be generalizable. 
Altogether, since this component seems to be heavily dependent upon its defining parameters, it may necessitate the operation of hyper-parameter tuning (e.g., by \texttt{irace} \cite{IRACE} or \texttt{SMAC} \cite{smac-2017}).

%\begin{figure*}
%\centering
%\includegraphics[width=0.48\columnwidth]{1p1boxplot_sphere_THF.eps}
%\includegraphics[width=0.49\columnwidth]{r2d2boxplot_sphere_THF.eps}
% \caption{Statistical boxplots accounting for 30 runs of minimizing the Sphere function with a stagnation threshold of $\vartheta=10^{-4}$ \textbf{using (tc-2)}, starting at $N_i=10$, and leveling-up to $N_f=10^4$ : m-lev-1p1 [LEFT] versus m-lev-sepC [RIGHT], comprising all three upscale operators, (U-1)-(U-3). \label{fig:boxplot_sphere_THF}}
%\end{figure*}

\section{Black-Box Test-Bed: Quantum Control Systems} \label{sec:systems}
The simulated control problems under investigation lie in the field of QC \cite{Hersch93,Hersch00,Gerber07}.
Single-objective optimization of a QC system aims at maximizing a simulation-based observable by shaping a temporal control electric field, $E\left(t\right)$, which determines the dynamics of the quantum process as dictated by the Schr\"odinger equation:
\begin{equation}
\label{eq:schroedinger}
\begin{array}{l}
\displaystyle i\frac{\partial \psi}{\partial t}=(\mathcal{H}_0+V)\psi(t)\\
\displaystyle V=-\mu E(t)\cos(\Omega_0 t)~,
\end{array}
\end{equation}
where $\mathcal{H}_0$ is the field-free system Hamiltonian, $\mu$ is the dipole moment and $\Omega_0$ is the field carrier frequency. Given an observable operator, $\mathcal{O}$, with the propagated wave function $\psi$ solving Eq.~\ref{eq:schroedinger}, a quantum expectation value is then defined as:
\begin{equation}
 \mathcal{J} = \left< \psi  \left|\mathcal{O} \right| \psi \right> .
\end{equation}
In an experimental laboratory setup, closed-loop learning is carried out by spectral phase modulation of $\phi\left(\Omega\right)$, while adhering to a fixed spectral amplitude $A\left( \Omega \right)$; the envelope of the electric field is described by:
\begin{equation}
\label{eq:efield} 
E(t)=\mathbb{R} \left\{\int A\left(\Omega\right)\exp(i\phi\left(\Omega\right)) \exp\left(-i\Omega t\right) \ d\Omega \right\}.
\end{equation}
The decision variables in such experimental control processes are mapped onto the spectral phase function, $\phi(\Omega)$, \textbf{discretized at $n_{\ell}$ frequencies} $\{\Omega_i\}_{i=1}^{n_{\ell}}$ that are equidistantly spaced across the bandwidth:
\begin{equation}
\label{eq:phase}
\phi(\Omega) = \left(\phi(\Omega_1),\phi(\Omega_2),...,\phi(\Omega_{n_{\ell}})\right).
\end{equation}
In practice, the pulse shaping process is implemented by a so-called Spatial Light Modulator (SLM), which can vary the phase elements in Eq.~\ref{eq:phase}. The simulated voltage applied to the $\jmath^{th}$ pixel in the pulse shaper must therefore be adjusted to set the value of $\phi\left(\Omega_{\jmath}\right)$, $\jmath=1,\ldots,n_{\ell}$.
To best simulate this experimental setup, we consider $n_{\ell}$ individual pixels subject to rectangle-activation-functions, $\textrm{squ}(\nu)$, (ideally) sharply-defined and with no gaps between each other. This is referred to as the \emph{staircase approximation}. 
\textbf{Elaboration on this so-called pixelation effect is provided within the Appendix (\ref{app:pixelation})}.
Practically, step-function gaps between SLM pixels are responsible for the construction of so-called \emph{parasitic replica pulses} in the temporal domain, which are located at the zeros of the {\bf sinc} envelope function.
Notably, strong phase variations from one pixel to another generally cause more pronounced replica pulses, which generally result in lower, suboptimal yields \cite{MatthiasPHD}. 
This effect is a consequence of fine \textit{discretization}, which is a central aspect of the current work, combined with invariance properties due to the shaping process (i.e., the Fourier transform in Eq.~\ref{eq:efield}). 
This effect can be responsible for the distortion of relatively simple QC objective functions (e.g., two-photon-absorption systems, which are considered in what follows). 
Importantly, nominally easy landscapes to be globally optimized can become artificially constrained with local traps in practice, due to this effect.
%\medskip

As a test-bed for our m-lev ESs' operation, we consider three \textbf{simulation-based} QC systems:
\begin{compactenum}[(S-1)]
\item Two-photon absorption (TPA) processes through dispersive media
\item Rotational population transfer of a diatomic molecule (starting from the $J=0$ rotational ground state)
\item Field-free molecular alignment of a diatomic molecule (at zero Kelvin, starting from $J=0$).
\end{compactenum}

\medskip

\textbf{The primary goal of utilizing this test-bed is to validate the efficacy of the m-lev approaches and compare it to the direct variants that operate at the finest scales.}

\subsection{TPA through Dispersive Media}
We consider effects of \textit{dispersion} on two-photon absorption processes (frequency doubling).
\textit{Experimental global optimization} of such a system in a dispersive toluene medium, driven by the CMA-ES heuristic, both single- and multi-objective, has already been accomplished (per $N_{d}=128$) as was reported in \cite{LaForgeMO}. Here, we consider its \textit{simulated} objective function \cite{GECCO07_SHG}.

To account for the effects of linear dispersion on the electric field, a propagation function is formulated by means of a \textit{fourth-order polynomial} to be discretized at the same frequencies:
\begin{equation}
k\left( \Omega \right) = k_2 \Omega^2 + k_3 \Omega^3 + k_4 \Omega^4,
\end{equation} 
constituting a Taylor expansion for nonabsorptive medium that discards zeroth- and first-order terms due to the invariance properties of the Fourier transform. Then, given a control phase $\phi(\Omega)$ and a dispersion term $k\left( \Omega \right)$, the spectral intensity $I_2(\Omega)$ is formulated by means of the spectral field terms: 
\begin{equation}
\label{eq:ewfield}
\begin{array}{l}
\displaystyle E_1(\Omega)=A(\Omega)\exp i\left[\phi(\Omega)+k(\Omega)\right]\\
\displaystyle E_2(\Omega)=E_1(\Omega)*E_1(\Omega)=
\int_{-\infty}^{\infty} E_1(\Omega') \cdot E_1(\Omega-\Omega') \textrm{d}\Omega'\\
\displaystyle I_2(\Omega)=|E_2(\Omega)|^2.
\end{array}
\end{equation}
The standard TPA objective function is subject to maximization and defined as follows:
\begin{equation}
\label{eq:SHG_T} \displaystyle f_{\tau} \left(\phi,k \right) = \int_{-\infty}^{\infty} I_2(\Omega) \textrm{d}\Omega \longrightarrow \max .
\end{equation}
Since $f_{\tau}$ is maximized in non-dispersive processes by any phase function linear in the frequency $\Omega$ \cite{GECCO07_SHG}, 
$$ \textrm{argmax}_{\phi\left(\Omega\right)}\left\{f_{\tau} \left(\phi,k=0 \right)\right\}\equiv a \Omega + b ~~~ \forall a,b $$
(and in particular by a constant phase, $a=0$, $\forall b$), the decision variables in the current dispersive process are meant to compensate for the function $k(\Omega)$ over the \textbf{periodic domain} $\left[0,2\pi\right]^{n_{\ell}}$.

We target this problem with the following dispersion profiles:
\begin{compactenum}[({TPA}-1)]
\setcounter{enumi}{-1}
\item $\hat{k}_0\left( \Omega \right):\left\{ k_2=0,~~k_3= 0,~~k_4=0 \right\}$
\item $\hat{k}_1\left( \Omega \right):\left\{ k_2=5000,~~k_3= 0,~~k_4=0 \right\}$
\item $\hat{k}_2\left( \Omega \right):\left\{ k_2=11300,~~k_3= 7990,~~k_4=2530 \right\}$
\item $\hat{k}_3\left( \Omega \right):\left\{ k_2=50000,~~k_3= 25000,~~k_4=10500 \right\}$
\end{compactenum}
Since the global maximum is known, by setting the decision variables to the compensating phase and calculating the objective function value, $f_{\tau,\hat{k}}^{\max}$, the problem is then transformed into a \textit{normalized} maximization problem with a \textbf{globally maximal unit value}:
\begin{equation}
\label{eq:maxTPA} \displaystyle f_{\textrm{TPA}} \left(\phi,\hat{k} \right) = \frac{f_{\tau} \left(\phi,\hat{k}\right)}{f_{\tau,\hat{k}}^{\max}} \longrightarrow \max .
\end{equation}
Despite some symmetry properties of the standard TPA function, it captures the complexity of the Fourier transform between the decision space to the evaluation space, and does not constitute a separable function \cite{GECCO07_SHG}. 

\subsection{Rotational Population Transfer}\label{sec:defRot}
The computational models for calculating rotational population transfer and molecular alignment were previously described \cite{Vrakking}, and were further considered for simulation-based single-objective \cite{Shir-JPhysB} and multi-objective \cite{Shir-MOQC} optimization.
Two \emph{electronic states} are taken into account, the ground $\left|g\right>$ and off-resonant excited electronic state $\left|e\right>$, with initial occupation of the ground $J=m=0$ rotational state. The wave function is expanded in the following fashion:
\begin{equation}
\label{expansion} 
\displaystyle \Psi(t) = \sum_{J=0}^{N_{rot}} \left[ \alpha_{J}^{(g)}(t) \psi_{g,J} + \alpha_{J}^{(e)}(t)\psi_{e,J} \right].
\end{equation}
We consider $N_{rot}=20$, where this expansion was confirmed to give converged results in the present calculations. 
The eigenvalues of $\mathcal{H}_0$ (i.e., a diagonal matrix) in Eq.~\ref{eq:schroedinger} are $E(J)=B_{rot}J(J+1)$ where $B_{rot}$ is referred to as the \textit{rotational constant}. 
The expansion in Eq.~\ref{expansion} is utilized in Eq.~\ref{eq:schroedinger}.
The (diatomic) molecule under investigation has a rotational constant of $B_{rot}=5\rm{cm}^{-1}$.
The transitions between $\left|g\right>$ and $\left|e\right>$ are assumed to proceed via the selection rules of the quantum numbers $\Delta J=1,~\Delta m=0$.
The peak Rabi frequency, which plays the role of specifying the laser intensity, is set to $\Omega_{ge}=160 \times 10^{12}\rm{s}^{-1}$.

We consider here the problem of population transfer within the rotational framework as an optimization problem. The objective to be maximized is defined as the probability to populate the $J_{target}=4$ rotational level, $\left|4\right>:=\left|gJ_{target}=4\right>$, upon starting in the initial ground state $\left|0\right>:=\left|gJ=0\right>$:
\begin{equation*}
\mathcal{J}:=\mathbf{Pr}\left\{\left|0\right>\leadsto \left|4\right>\right\}.
\end{equation*}
The formal objective function value, at an observation time $t=t_{\textrm{measure}}$, reads:
\begin{equation}
\label{rotObjective}
\displaystyle f_{\textrm{Rot}} = \left|\alpha_{J_{target}=4}^{(g)}(t_{\textrm{measure}})\right|^2  \longrightarrow \max .
\end{equation}

\subsection{Molecular Alignment}
The numerical modeling of rotational motion, presented in the previous section, is adopted here to the molecular alignment problem as well. 
We focus on a simplified variant of the standard alignment problem, starting at zero temperature ($T=0^{\circ}~K$) and with the initial state being $J=m=0$.
The same characteristic parameters of the rotational population transfer problem are set here, including the peak Rabi frequency as specified in Section~\ref{sec:defRot}.
The definition of the \emph{alignment observable} corresponds to $\cos ^2 \theta$, where $\theta$ is the angle of the internuclear axis with respect to the control field's polarization axis. Consequently, the objective function is defined as follows:
\begin{equation}
\label{alignObjective} 
\displaystyle f_{\textrm{Align}}=\rm{max}_{E(t)} \left<\cos^2(\theta)\right> \longrightarrow \max.
\end{equation}

\subsection{Implementation and Simulation Considerations}
The simulations consider a femtosecond laser pulse for the three QC systems, centered at $795\textrm{nm}$ with a bandwidth of $\Delta\lambda\sim {28}\textrm{nm}$, to be delivered to a pulse shaper with a programmable $n_{\ell}$ pixel SLM for phase-only modulation. 
%which gives pulses of duration $\Delta \tau \sim 2\textrm{fs}$ full-width-half-maximum (FWHM).
The search is to be conducted in the dimensionality range of $N_i=16,~N_f=2^{14}=16,384$, except for the alignment (S-3) problem which starts at $N_i=64$ (it is explained by the need to shape a more complex pulse-train \cite{Shir-JPhysB}).
Initial candidate solutions are to be randomly and uniformly generated within $\left[0,2\pi \right]^{N_i}$.
All simulation-based optimization problems (S-1)-(S-3) are transformed into minimization problems. The initial global step-size is set to $\sigma_0=\frac{2\pi}{3}$. 
Accounting for the periodic boundary conditions, a wrapping operator \cite{LabES} is applied to the decision vectors immediately after the mutation, i.e., $\phi\left(\Omega\right)\leftarrow \phi\left(\Omega\right) \textrm{mod}2\pi$.
All three upscale operators, (U-1)-(U-3), are tested per each instantiation to yield altogether six m-lev variants.
The m-lev-sepC starts with a $(6_W,12)$ strategy on $N_i$ and concludes with a $(16_W,33)$ strategy for the full-scale problem on $N_f$.
For the overall budget of objective function calls, a total of $3\cdot 10^5$ evaluations is granted per each run. 

Regarding termination criteria, \textit{fixed-target} optimization by (tc-1) is a natural choice due to the available normalization of all three objective functions, and the intuitive interpretation of $\epsilon$. 
Nevertheless, we applied preliminary attempts to obtain a stable (tc-2) implementation over $f_{\textrm{TPA-0}}$ , which demanded extensive efforts of hyper-parameter tuning. Eventually, while the (tc-2)-based m-lev-1p1 variants performed similarly to (tc-1), the (tc-2)-based m-lev-sepC variants performed much worse than (tc-1), up to an increased order of magnitude in the utilized evaluations.
\textbf{In practice, all runs to be reported in Section \ref{sec:observations} were conducted using (tc-1) with a fixed-target of $\epsilon=0.05$}. 
%of (S-1) and $\cdot 10^5$ evaluations per each run of (S-2) and (S-3).

%\textcolor{armygreen}{The direct variants are to be run on a scale of $2^{10}$ decision variables.}
The \textit{direct} variants are run on scales, as high as possible, which enable reasonable convergence and overall computation time.
Note that the computation time of a single shaping simulation call dramatically increases with the dimensionality $n_{\ell}$, especially due to the required FFT procedure \cite{NumericalRecipes}; the latter performs \textit{de facto} in $\mathcal{O}(n)$ complexity, e.g., resulting in a duration of more than $6\texttt{sec}$ for a single TPA function call featuring $2^{14}$ decision variables on a state-of-the-art workstation (Dual Intel Xeon Processor E5-2670 v3).
Additionally, the elapsed CPU time for simulating (S-2) and (S-3) is roughly $5\texttt{sec}$ for a single electric field.
Therefore, the incentive to apply multileveling to this class of problems is twofold -- a smaller number of total simulation calls, but at the same time, less fine-scale simulation calls. In fact, targeting this class of problems by a direct ES over $2^{10}$ decision variables already becomes an impractical computational task that requires several days per each run using a state-of-the-art workstation.

\section{Simulation-Based Observations} \label{sec:observations}
Tables \ref{table:experimentS1} and \ref{table:experimentS2S3} present the statistics of the utilized simulation (evaluation) calls to reach $\epsilon=0.05$ away from the global optimum, over 30 runs, for the six m-lev ES instantiations when targeting the three systems (an exception is (S-3) due to its excessive computation time).
In what follows, we independently examine the test-cases in detail and conduct the aforementioned statistical protocol of a Friedman test (ensuring that at least one variant has significant differences in performance) followed by Mann-Whitney U-tests, considering the statistical hypotheses $h_0$ and $h_1$ as stated in Section~\ref{sec:spherePOC}.
\begin{table*}
%\centering Experimental Observation\\
\begin{scriptsize}
\begin{tabular}{| l | l | l | l | l |}
\hline
ES / (S-1)-variant & $f_{\textrm{TPA-0}}$ & $f_{\textrm{TPA-1}}$ & $f_{\textrm{TPA-2}}$& $f_{\textrm{TPA-3}}$ \\
\hline
\hline
m-lev-1p1: (U-1) & \textbf{4248.93 $\pm$2025.56 (3966.5)} & \textbf{2928.6 $\pm$1799.48 (957)} & \textbf{5639.87 $\pm$2030.74 (4398)} & \textbf{15423.9 $\pm$2419.53 (13255)} \\
\hline
m-lev-1p1: (U-2) & 19520.2 $\pm$5159.73 (21853) & 27004.1 $\pm$7270.7 (25084.5) & 24734.5 $\pm$3508.5 (24589.5) & 67143.3 $\pm$19952.2 (50641.5) \\
\hline
m-lev-1p1: (U-3) & 14710 $\pm$5122.27 (14622.5) & 15353.8 $\pm$4084.38 (15905) & 19267.1 $\pm$3255.94 (17803.5) & 52987.7 $\pm$14513 (46169.5) \\
\hline
 \hline
% & & & \\
m-lev-sepC: (U-1) & \textbf{2824.73 $\pm$1310.64 (1454)} & \textbf{11774.5 $\pm$24384.3 (1573)} & \textbf{3213.27 $\pm$628.234 (2770)} & \textbf{15515 $\pm$1633.66 (15314)}\\
\hline
m-lev-sepC: (U-2) & 11186.9 $\pm$5203.14 (9939.5) & 11359 $\pm$3189.13 (10427) & 17252.7 $\pm$2337.42 (16956.5) & 55675.9 $\pm$9390.88 (50437) \\
\hline
m-lev-sepC: (U-3) & 8598.9 $\pm$1710.38 (8904) & 11512.7 $\pm$6076.62 (8715) & 11752.7 $\pm$1438.1 (12634.5) & 41487.2 $\pm$3220.31 (39874.5)\\
\hline
\end{tabular}
\end{scriptsize}
\caption{Mean with t-distribution confidence intervals at the 99\% level, as well as median values (in parentheses), of the required evaluation calls to reach the threshold, over 30 runs, for the six m-lev ES instantiations on \textbf{(S-1)}. 
Bold entries indicate statistically-significant best performance of the utilized operator per each heuristic (m-lev-1p1, m-lev-sepC). 
On this problem, utilization of (U-1) performed best, as elaborated in the text. \label{table:experimentS1}}
\end{table*}
\begin{table*}
%\centering Experimental Observation\\
\begin{footnotesize}
\begin{tabular}{| l | l | l |}
\hline
ES-variant/System & $f_{\textrm{Rot}}$ (S-2) & $f_{\textrm{Align}}$ (S-3) \\
\hline
\hline
m-lev-1p1: (U-1) & \textbf{17505.9 $\pm$18890.2 (724.5)} & 206820$\pm$ 63623.4 (300000) [s.r.: 11] \\
\hline
m-lev-1p1: (U-2) & 11723.7 $\pm$15084.5 (1401.5)& 190375$\pm$ 65096.5 (300000) [s.r.: 13] \\
\hline
m-lev-1p1: (U-3) & 5386.57 $\pm$9028.74 (1843) & 255081$\pm$ 45505.7 (300000) [s.r.: 7] \\
\hline
 \hline
m-lev-sepC: (U-1) & \textbf{1510.93 $\pm$394.46 (1353)} & 170958 $\pm$66666.8 (199503) [s.r.: 15]\\
\hline
m-lev-sepC: (U-2) & 2431.5 $\pm$360.33 (2352) & 210525 $\pm$59172.8 (300000) [s.r.: 12] \\
\hline
m-lev-sepC: (U-3) & 8664.73 $\pm$12498.5 (2163) & 190014 $\pm$58004.9 (232245) [s.r.: 16] \\
\hline
\end{tabular}
\end{footnotesize}
\caption{Mean with t-distribution confidence intervals at the 99\% level, as well as median values (in parentheses) and number of successful runs [s.r.], of the required evaluation calls to reach the threshold, over 30 runs, for all six m-lev ES instantiations on \textbf{(S-2)} and only for three m-lev-sepC utilizations on \textbf{(S-3)}. 
Bold entries indicate statistically-significant best performance of the utilized operator per each heuristic (m-lev-1p1, m-lev-sepC). 
Utilization of (U-1) performed best on (S-2), whereas all operators performed equally well on (S-3) | as elaborated in the text.
\label{table:experimentS2S3}}
\end{table*}

\subsection{TPA (S-1)}
\begin{figure*}
\centering
\includegraphics[width=0.99\columnwidth]{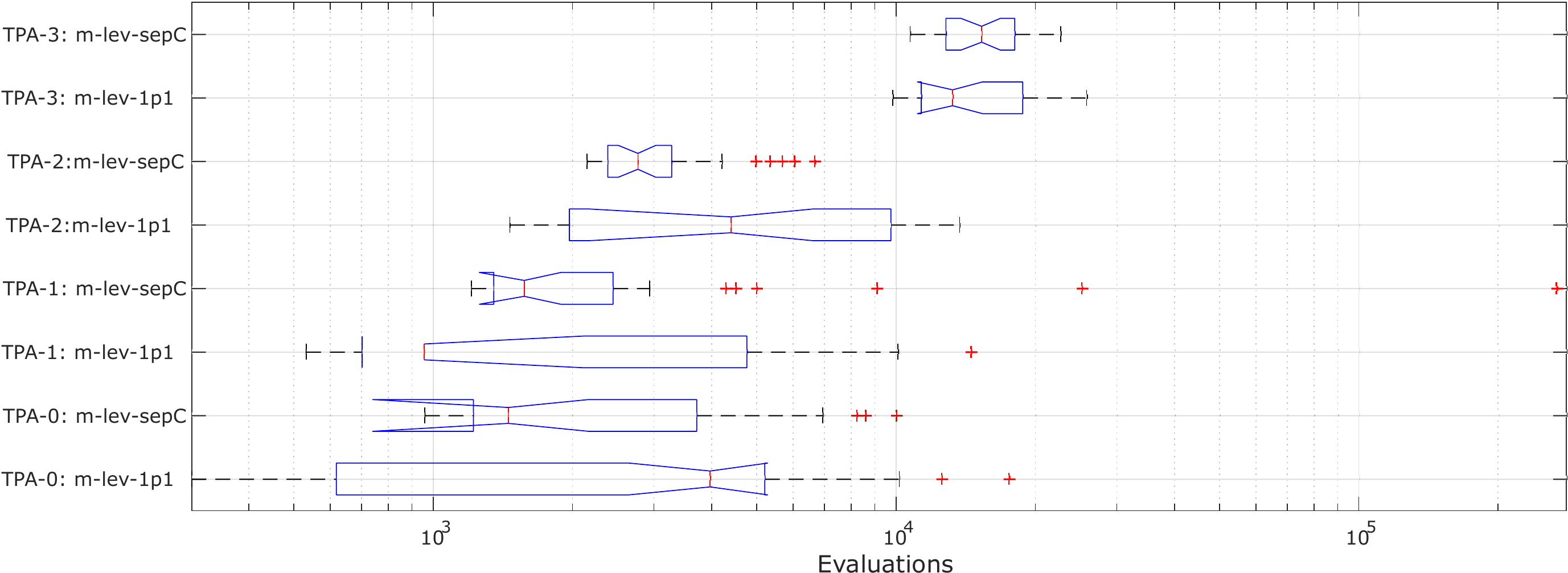}
 \caption{Statistical boxplots accounting for 30 runs of maximizing the four TPA systems (S-1) starting at $N_i=16$, leveling-up to $N_f=2^{14}$, \textbf{utilizing (U-1)}, and \textbf{using (tc-1)} with a target of $\epsilon=0.05$: m-lev-1p1 and m-lev-sepC are depicted vertically adjacent per each TPA problem-instance. \label{fig:boxplot_S1U1}}
\end{figure*}

\begin{figure*}
\centering
\includegraphics[width=0.49\columnwidth]{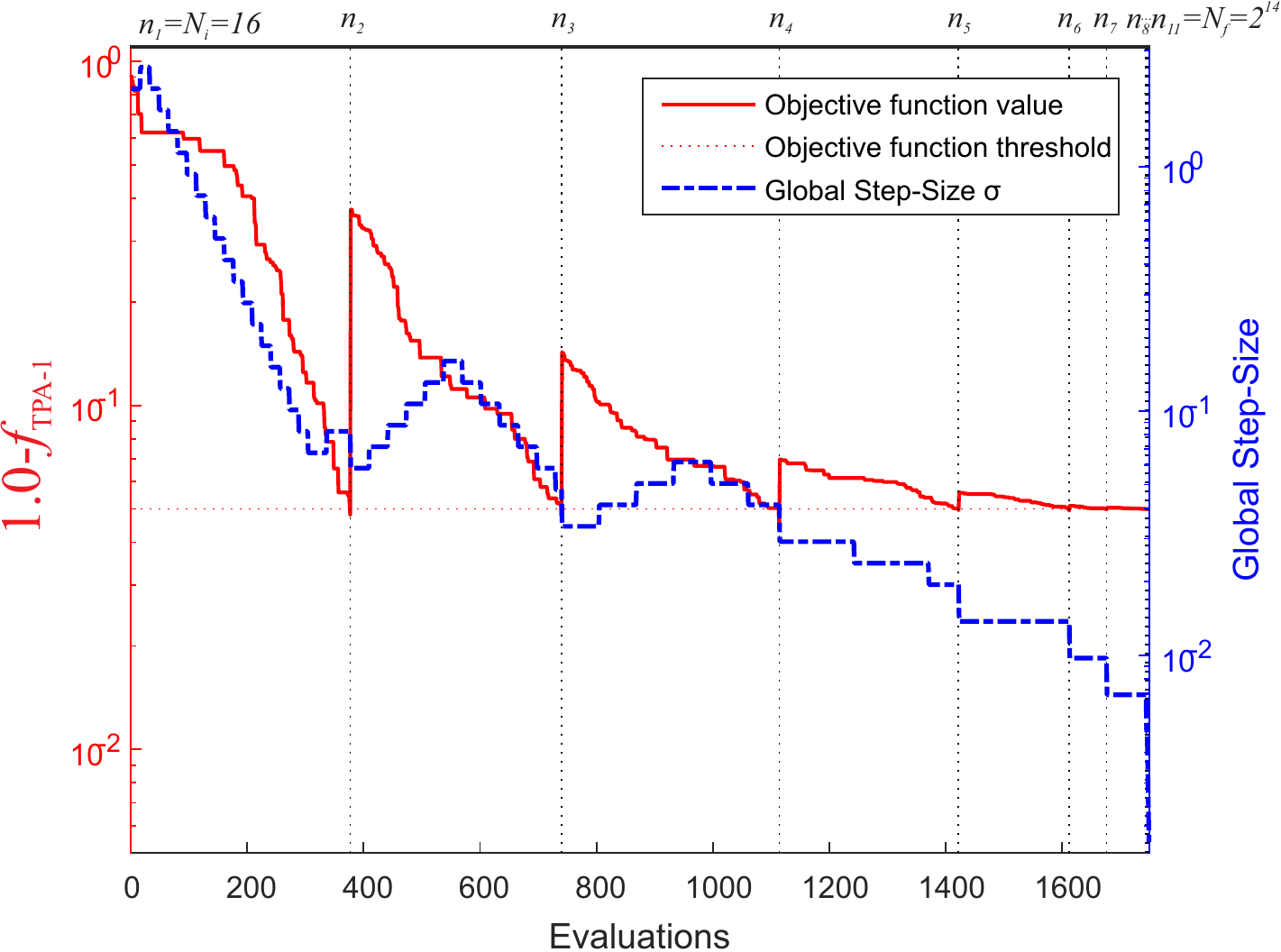}
\includegraphics[width=0.49\columnwidth]{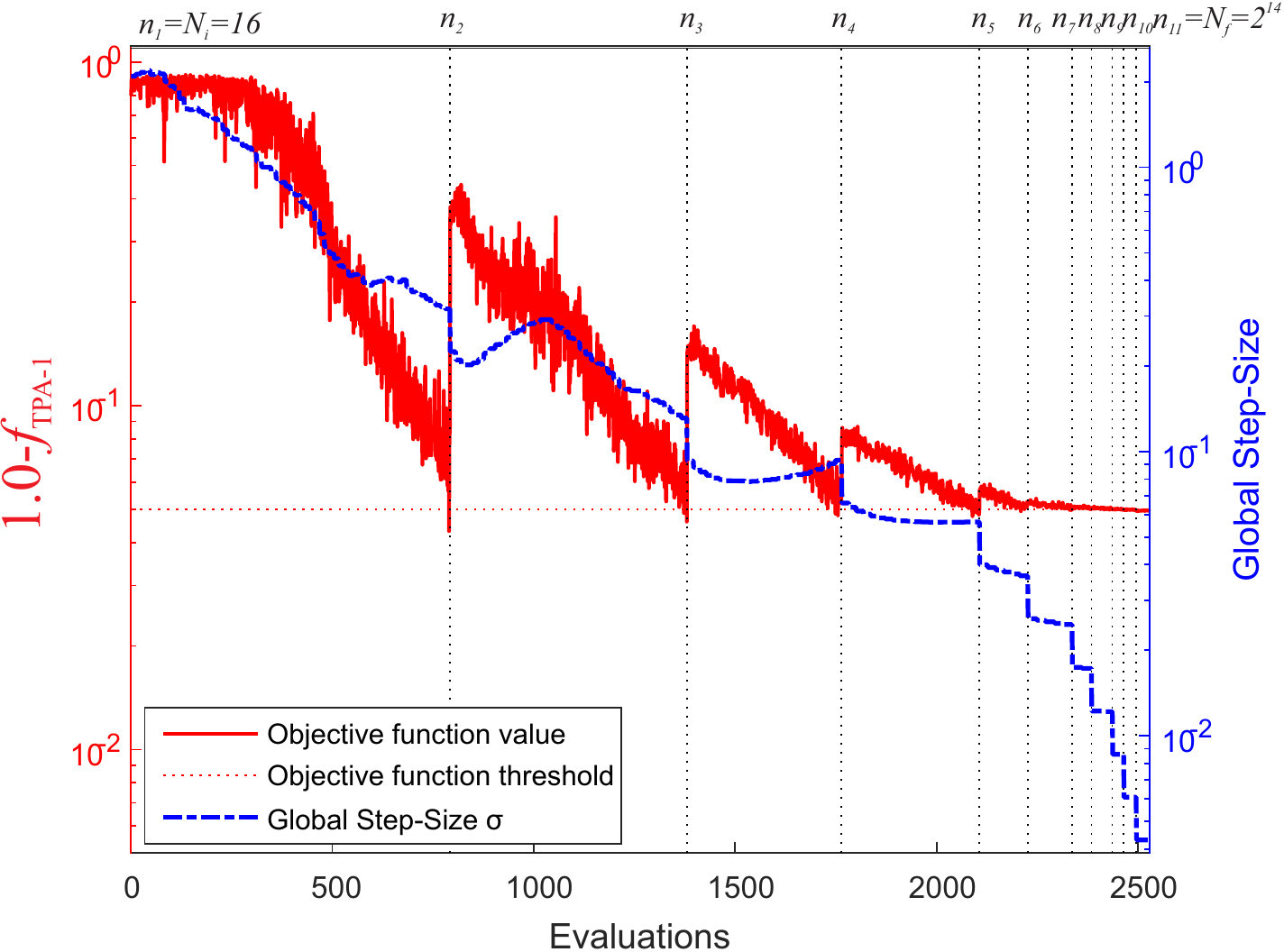}
 \caption{Selected runs of m-lev-1p1 [LEFT, 65\%-tile run] and m-lev-sepC [RIGHT, 75\%-tile run] applied to $f_{\textrm{TPA-1}}$ with $N_f=2^{14}$, using (tc-1) with a target of $\epsilon=0.05$, both employing (U-1). Each run starts at $N_i=16$, with vertical dashed lines that represent each leveling, corresponding to the displayed $n_{\ell}$ values (top x-axis). The objective function values (left y-axis, depicting $1.0-f_{\textrm{TPA-1}}$) and global step-size (right y-axis) are both displayed on log-scales as a function of the utilized objective function evaluations. \label{fig:ml_TPA}}
\end{figure*}

Both m-lev-1p1 and m-lev-sepC operated best on (S-1) when utilizing (U-1), by a clear margin.  
The statistical comparisons using a Friedman test, followed by Mann-Whitney U-tests concluded that $h_0$ was always rejected on both m-lev variants. 
Thus, (U-1) significantly performs better on this problem. 
Statistical comparisons between m-lev-1p1 and m-lev-sepC show that they performed equally well when utilizing (U-1) on the TPA case, except for $f_{\textrm{TPA-1}}$, where m-lev-1p1 outperformed m-lev-sepC ($h_0$ was rejected on the latter at the 5\% significance level). 
Fig.~\ref{fig:boxplot_S1U1} provides the statistical boxplots reflecting the 30 runs of both m-lev instantiations when utilizing (U-1) on the TPA systems (S-1).  
We note the large confidence intervals of the number of evaluations used to maximize these functions considering the different runs.
We also mention the numerical observation from Table \ref{table:experimentS1} that m-lev-1p1 optimized $f_{\textrm{TPA-1}}$ more easily than $f_{\textrm{TPA-0}}$ when utilizing (U-1), which is surprising to some extent. A statistical comparison ruled out significant outperformance though, and yet we offer the following explanation for this behavior. 
The m-lev-1p1 is prone to getting trapped in local optima (due to its elitist selection mechanism), which are in practice more prominent in the staircase approximation for non-dispersive TPA in comparison to the lightly-dispersive TPA. 
Fig.~\ref{fig:ml_TPA} depicts selected runs of m-lev-1p1 and m-lev-sepC on $f_{\textrm{TPA-1}}$.
The \textit{direct} ESs could not converge within a practical duration on scales larger than $2^{10}$. 
We ran $(1+1)$-ES and $(\mu_W,\lambda)$-sep-CMA-ES to maximize $f_{\textrm{TPA-1}}$ on a scale of $2^{10}=1024$ variables. Although the numerical results are not comparable to the m-lev variants, we note that the $(1+1)$-ES reached the threshold within an average of 40504.5$\pm$ 1138.87 function evaluations (median was 40525), whereas the direct $(12_W,24)$-sep-CMA-ES reached it within an average of 88485.6$\pm$ 11370.5 evaluations (median was 82176).
To summarize, the practical proficiency of the m-lev framework is twofold: (1) solving a high-resolution QC problem which could not be solved in its full-scale formulation within a practical run-time, and (2) an order of magnitude improvement when applied to $f_{\textrm{TPA-1}}$.

\subsection{Rotational Population Transfer (S-2)} 
\begin{figure*}
\centering
\includegraphics[width=0.99\columnwidth]{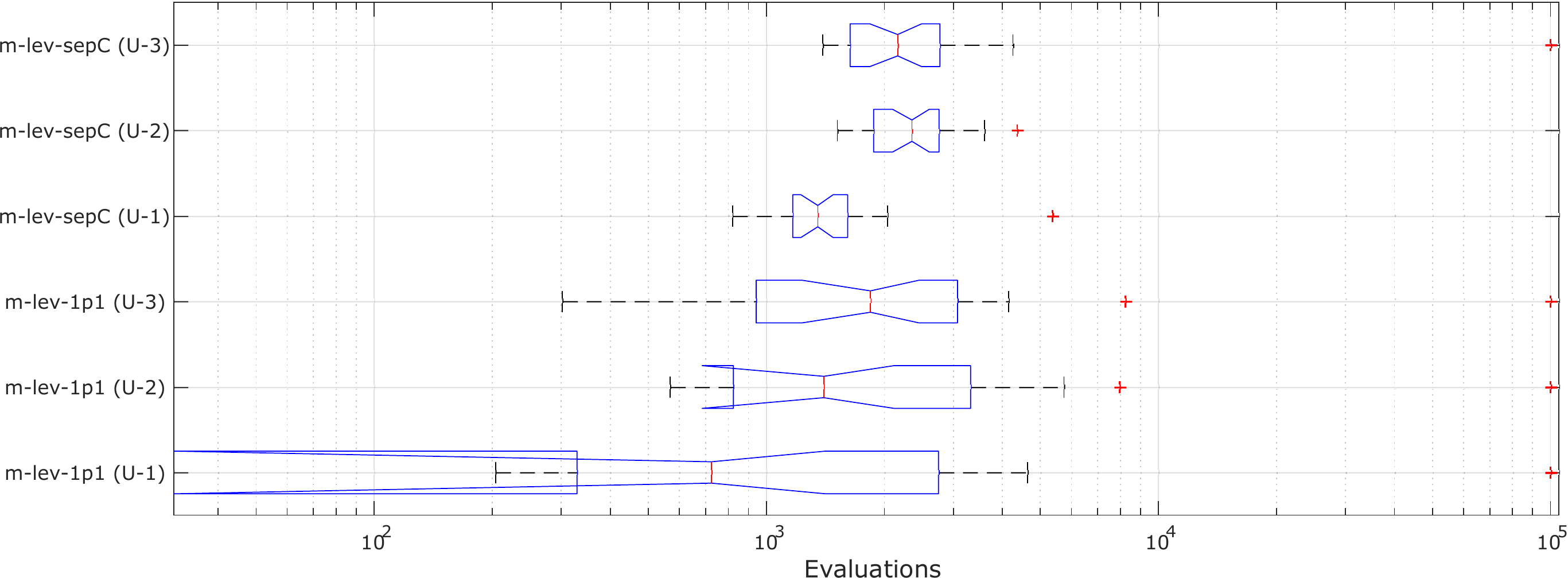}
\caption{Statistical boxplots accounting for 30 runs of maximizing $f_{\textrm{Rot}}$ (S-2) starting at $N_i=16$, leveling-up to $N_f=2^{14}$, \textbf{using (tc-1)} with a target of $\epsilon=0.05$, and considering all three upscale operators: m-lev-1p1 [bottom three] versus m-lev-sepC [top three].\label{fig:boxplot_S2}}
\end{figure*}

\begin{figure*}
\centering
\includegraphics[width=0.48\columnwidth]{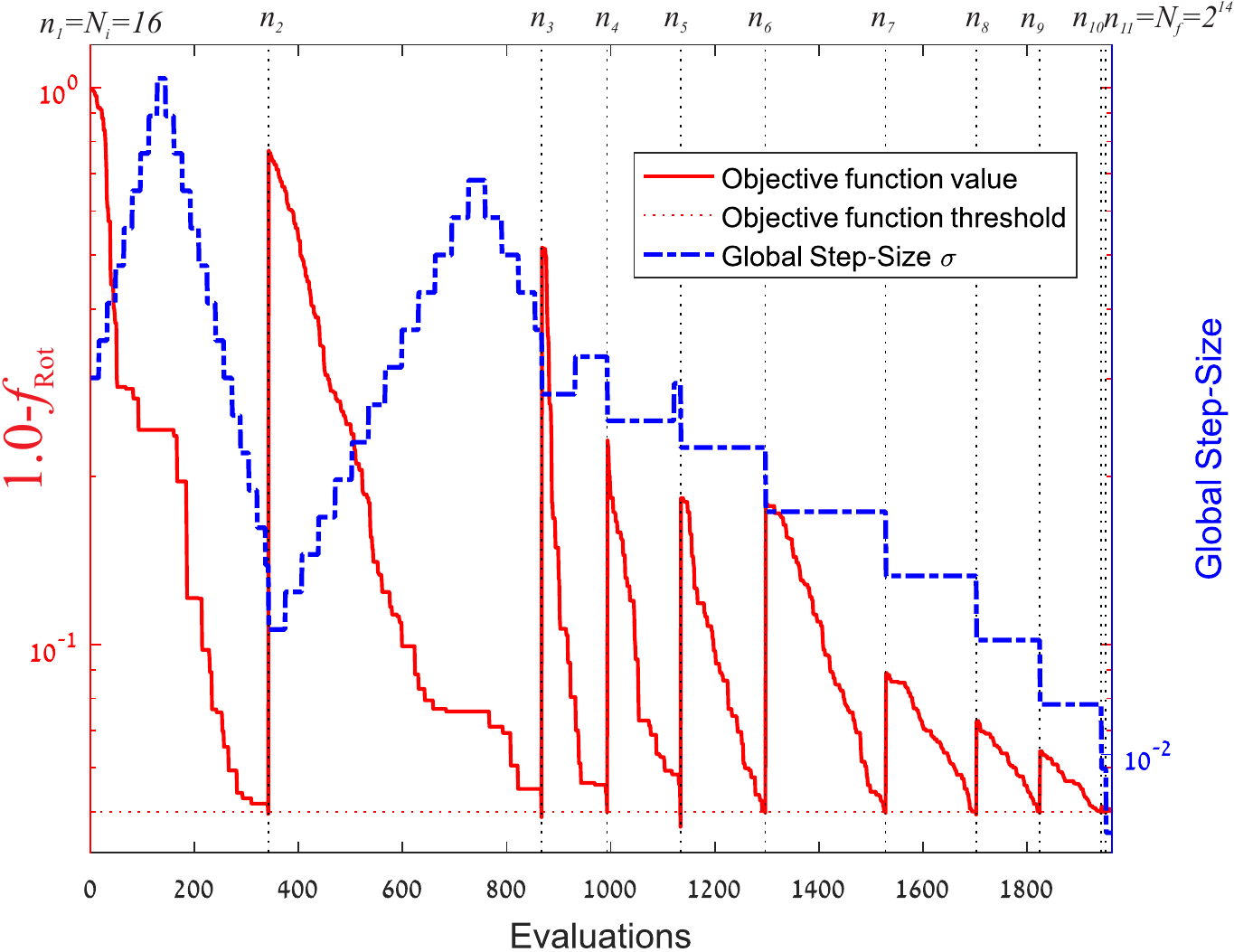} ~~~
\includegraphics[width=0.48\columnwidth]{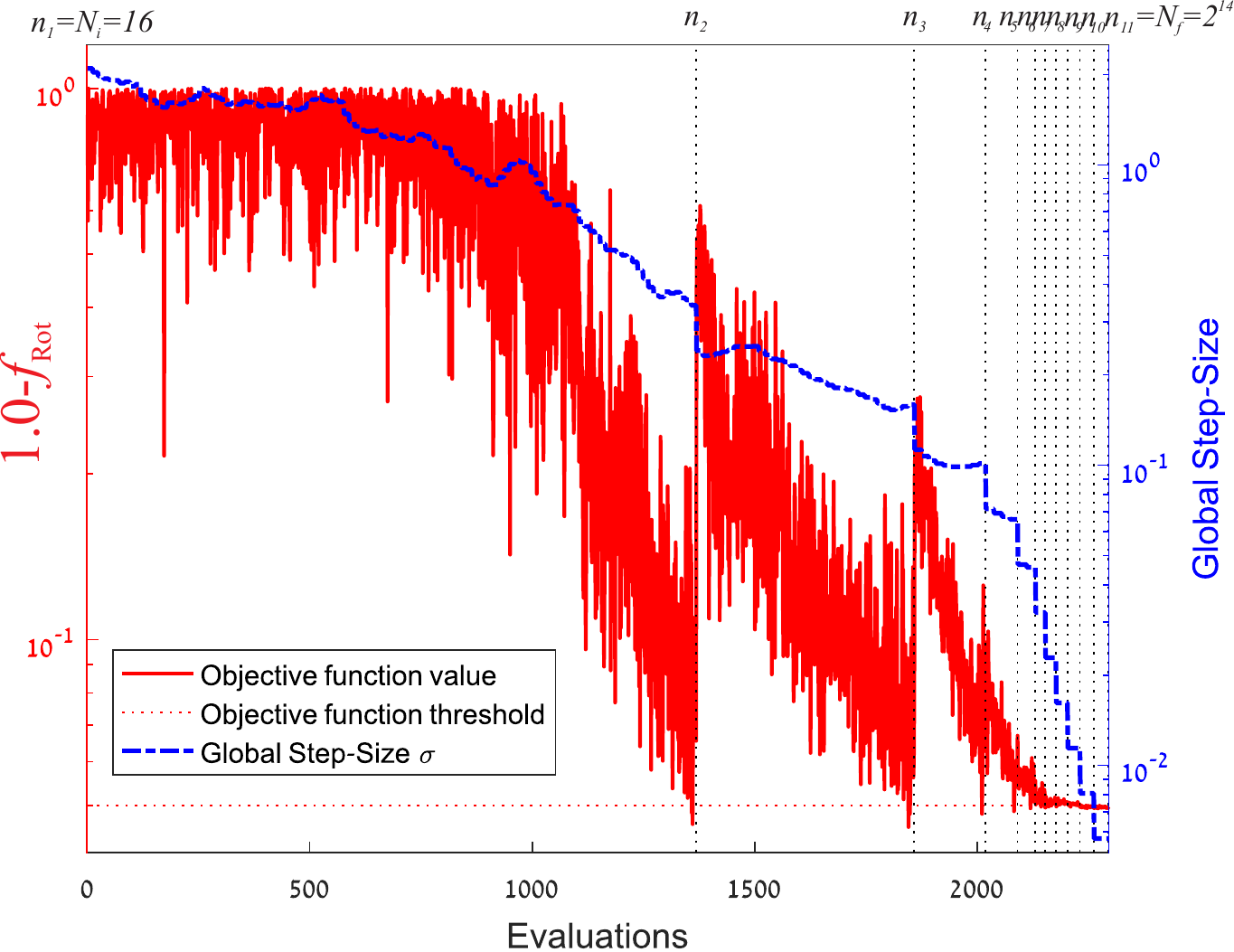}
\caption{Median runs of m-lev-1p1 [LEFT] and m-lev-sepC [RIGHT] applied to $f_{\textrm{Rot}}$ \textbf{(S-2)} with $N_f=2^{14}$, using (tc-1) with a target of $\epsilon=0.05$, employing (U-3) and (U-2), respectively. 
Each run starts at $N_i=16$, with vertical dashed lines that represent each leveling, corresponding to the displayed $n_{\ell}$ values (top x-axis). The objective function values (left y-axis, depicting $1.0-f_{\textrm{Rot}}$) and global step-size (right y-axis) are both displayed on log-scales as a function of the utilized objective function evaluations. \label{fig:ml_Rot}}
\end{figure*}

In addition to Table \ref{table:experimentS2S3}, Fig.~\ref{fig:boxplot_S2} provides the statistical boxplots reflecting the 30 runs of each m-lev variant on $f_{\textrm{Rot}}$ (S-2).
It is evident that m-lev-1p1 did not converge within the budget of $10^5$ evaluations on multiple occasions (reflected by the outliers), likely due to its tendency to get trapped in local optima. m-lev-sepC did not converge on two occasions when utilizing (U-3).
The statistical tests indicate that both m-lev-1p1 and m-lev-sepC operated best on (S-2) when utilizing (U-1), while their performance using either (U-2) or (U-3) was equally good ($h_0$ could not be rejected). 
This observation is aligned with the results of the TPA runs.
Comparing the two ESs on (U-1), their performance significantly differs as $h_0$ was rejected. Although m-lev-1p1 required function evaluations at a median of approximately half the median of m-lev-sepC, it did not converge within the total budget on 5 out of the 30 runs. 
We therefore argue that m-lev-sepC is preferable for treating this use-case.
Fig.~\ref{fig:ml_Rot} depicts median runs of both m-lev-1p1 and m-lev-sepC on $f_{\textrm{Rot}}$. %Evidently, 

To test \textit{direct} behavior on a fine-scale, we ran $(12_W,24)$-sep-CMA-ES to maximize $f_{\textrm{Rot}}$ on a scale of $2^{10}=1024$ variables. It reached the threshold within an average of 13114.3 $\pm$696.3 function evaluations (median was 12900). Also in this test-case, the m-lev approach obtained an order of magnitude improvement (when considering m-lev-sepC).

\subsection{Molecular Alignment (S-3)}
\begin{figure*}
\centering
\includegraphics[width=0.48\columnwidth]{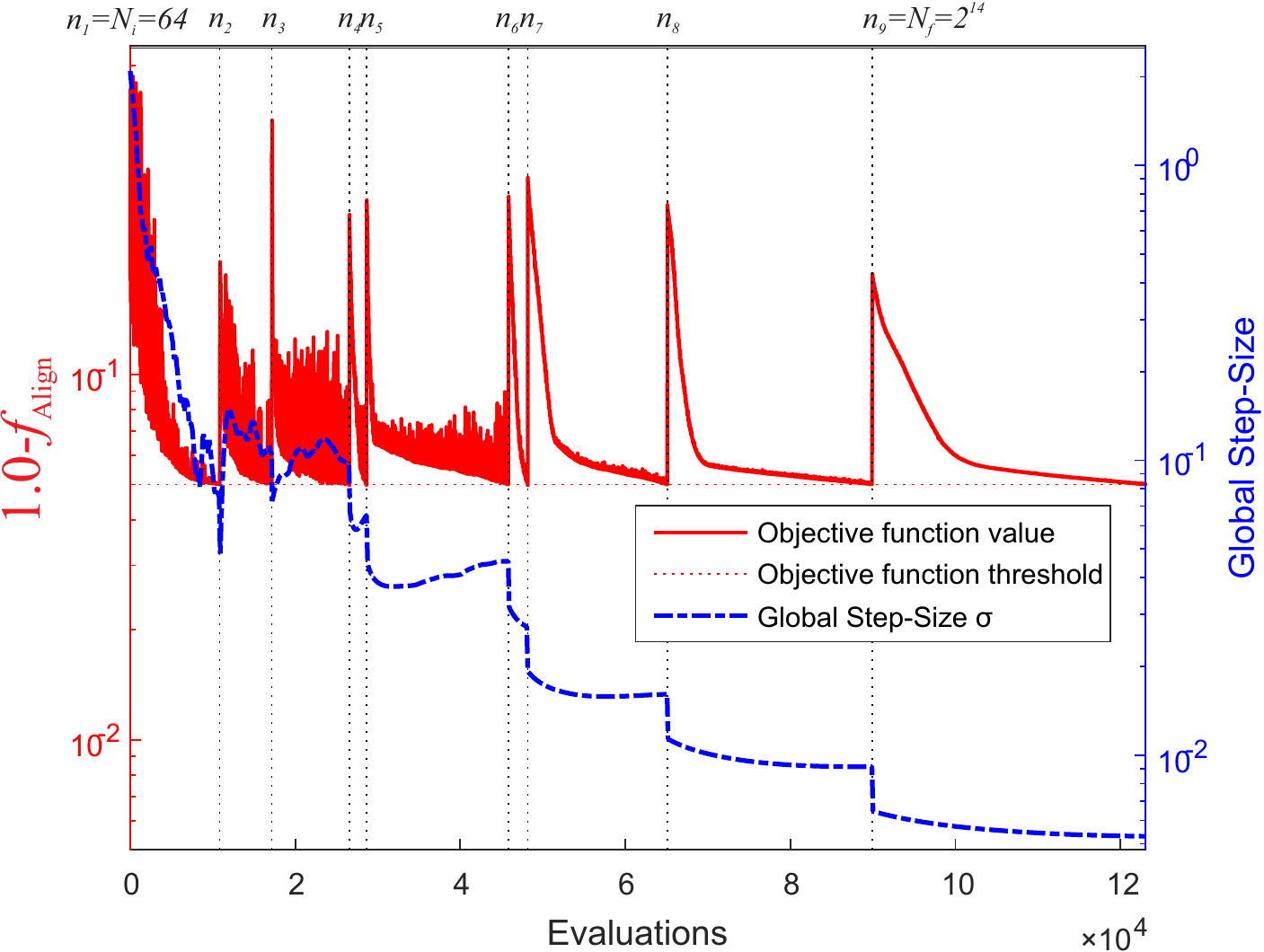} ~~~
\includegraphics[width=0.42\columnwidth]{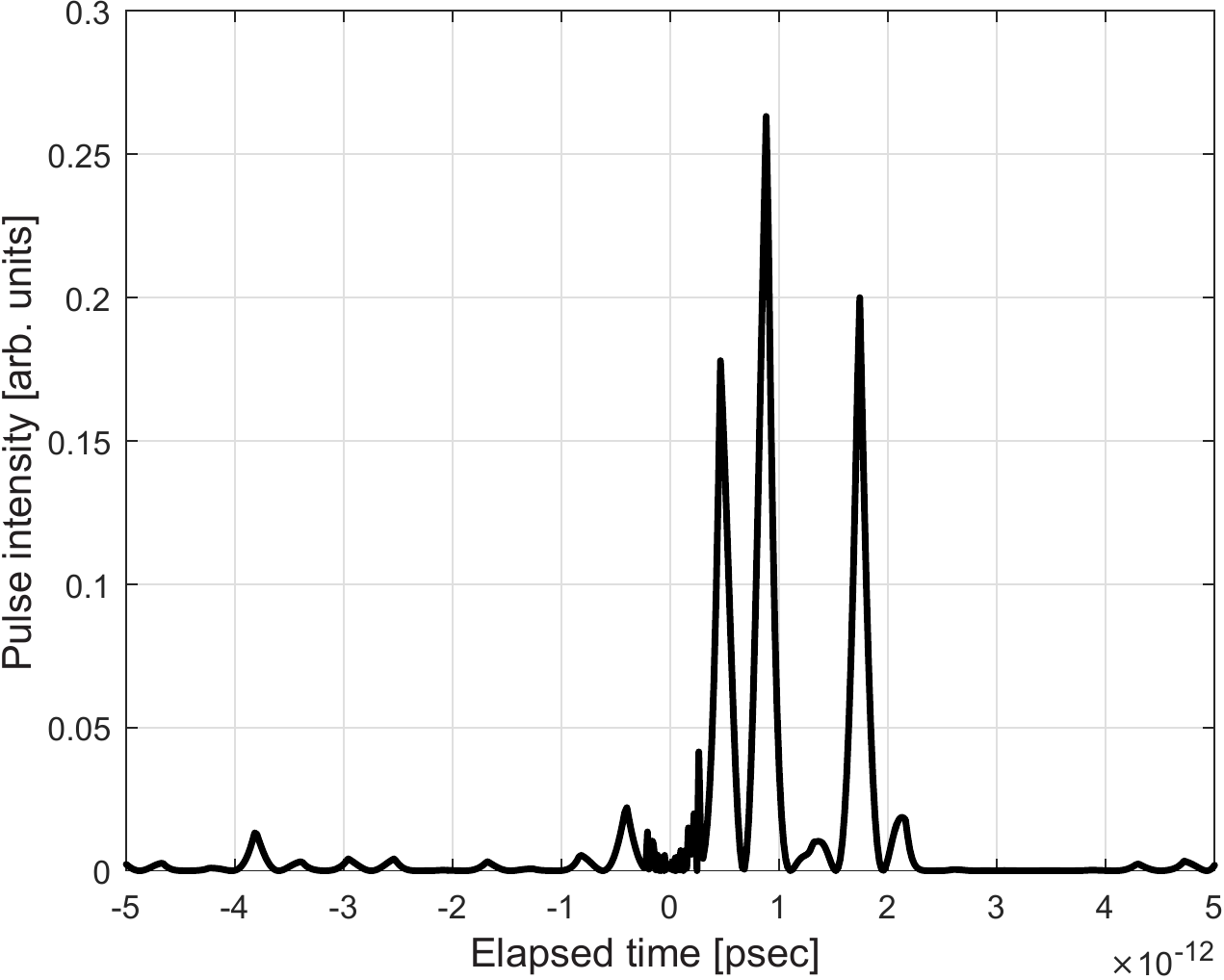}
 \caption{[LEFT] The median run of m-lev-sepC applied to $f_{\textrm{Align}}$ \textbf{(S-3)} with $N_f=2^{14}$, targeting a threshold of $\epsilon=0.05$ and employing (U-3). The run starts at $N_i=64$, with vertical dashed lines that represent each leveling, corresponding to the displayed $n_{\ell}$ values (top x-axis). The objective function values (left y-axis, depicting $1.0-f_{\textrm{Align}}$) and global step-size (right y-axis) are both displayed on log-scales as a function of the utilized objective function evaluations. 
[RIGHT] The attained pulse shape for the displayed run, obtaining a $\left<\cos^2(\theta)\right>$-alignment of 95\%.%This solution  
 \label{fig:ml_Align}}
\end{figure*}

Due to the excessive computation time and problem difficulty, we limit the total number of simulation calls to 300,000 per run. 
Table \ref{table:experimentS2S3} presents the statistics of the six m-lev instantiations on (S-3).
Evidently, this problem introduces a higher level of optimization difficulty, as the number of successful runs that meet the \textit{fixed-target} is low, typically less than half (this is also reflected in the
average number of function evaluations required to meet the target, in comparison to the other simulation-based QC systems).
The statistical tests indicate that m-lev-sepC performed equally well when utilizing (U-1), (U-2) or (U-3) ($h_0$ could not be rejected). For m-lev-1p1, (U-2) performed significantly better than (U-3), but equally well when compared to (U-1).
We argue that m-lev-sepC is preferable for treating this use-case, due to the higher rates of successful runs.
Fig.~\ref{fig:ml_Align} depicts the median run of m-lev-sepC on $f_{\textrm{Align}}$ as well as the attained solution's field intensity.

To test direct search on a fine-scale, we ran $(12_W,24)$-sep-CMA-ES to maximize $f_{\textrm{Align}}$ on a scale of $2^{10}=1024$ variables. 
It reached the threshold within an average of 296023$\pm$157075 function evaluations (median was 217644), which lies in the same order of magnitude as m-lev-sepC running on the finer-scale of $2^{14}$ variables, but is worst than m-lev-sepC (U-1).
%To test direct search on a fine-scale, we ran $(9_W,18)$-sep-CMA-ES to maximize $f_{\textrm{Align}}$ on a scale of $2^{7}=128$ variables. 
%It reached the threshold within an average of 125594$\pm$ 118277 function evaluations (median was 78966), which lies in the same order of magnitude as ML-sepC running on the fine-scale of $2^{14}$ variables.
%Counter-intuitively, direct search at $N=2^{11}=2048$ variables, with $(14_W,27)$-sep-CMA-ES, obtained better performance when it reached the threshold within an average of 58593$\pm$ 27909.9 function evaluations (median was 60034).

\section{Experimental Proof-of-Concept} \label{sec:experiments}
Unlike simulation-based optimization, which is broadly exercised throughout the sciences, direct experimental optimization is routinely deployed in QC experiments, but rarely elsewhere \cite{SeqExpEA_TutorialGECCO2018} -- due to the expenses involved.
In this section we describe the experimental results when the m-lev proposed approach was applied to the TPA use-case, and the laboratory setup details are provided in the Appendix (\ref{app:LabSetup}). %(\ref{app:LabSetup}).
Importantly, the experimental QC system is used for algorithmic validation purposes.

\subsection{Algorithmic Selection}
Due to the existence of noise and uncertainty in the experimental system, special attention is given to the strategy selection. 
As revealed by theoretical studies \cite{ArnoldNoise}, single-parent strategies experience difficulties in handling noisy landscapes, in comparison to multi-parent strategies: the application of recombination in the latter case proved highly efficient in
treating excessive noise, and it is typically attributed to Beyer's Genetic Repair Hypothesis \cite{Beyer97}.
More specifically, in the context of QC experimental optimization, the CMA-ES was observed in \cite{QCE_GECCO08,LabES} to fail without recombination, and to perform extremely well otherwise, as expected from theory.
At the same time, elitist strategies support the survival of parents, and are likely to encounter scenarios in which highly overvaluated objective function values are kept for long periods, causing stagnation (see, e.g., \cite{Beyer93}).
Following these considerations, we employ only the $(\mu_W,\lambda)$-sep-CMA-ES and m-lev-sepC strategies in the experiment.

Importantly, due to the lack of known \textit{a priori} normalization, the experimental optimization is run using (tc-0), that is, a ``fixed budget'' scheme. Each level is thus treated for a fixed number of iterations, yet with different number of measurements, since the population sizes vary.
Nearest-neighbor interpolation (U-1) is utilized.
\subsection{Experimental Results}
\begin{figure*}
\centering
\includegraphics[width=0.5\columnwidth]{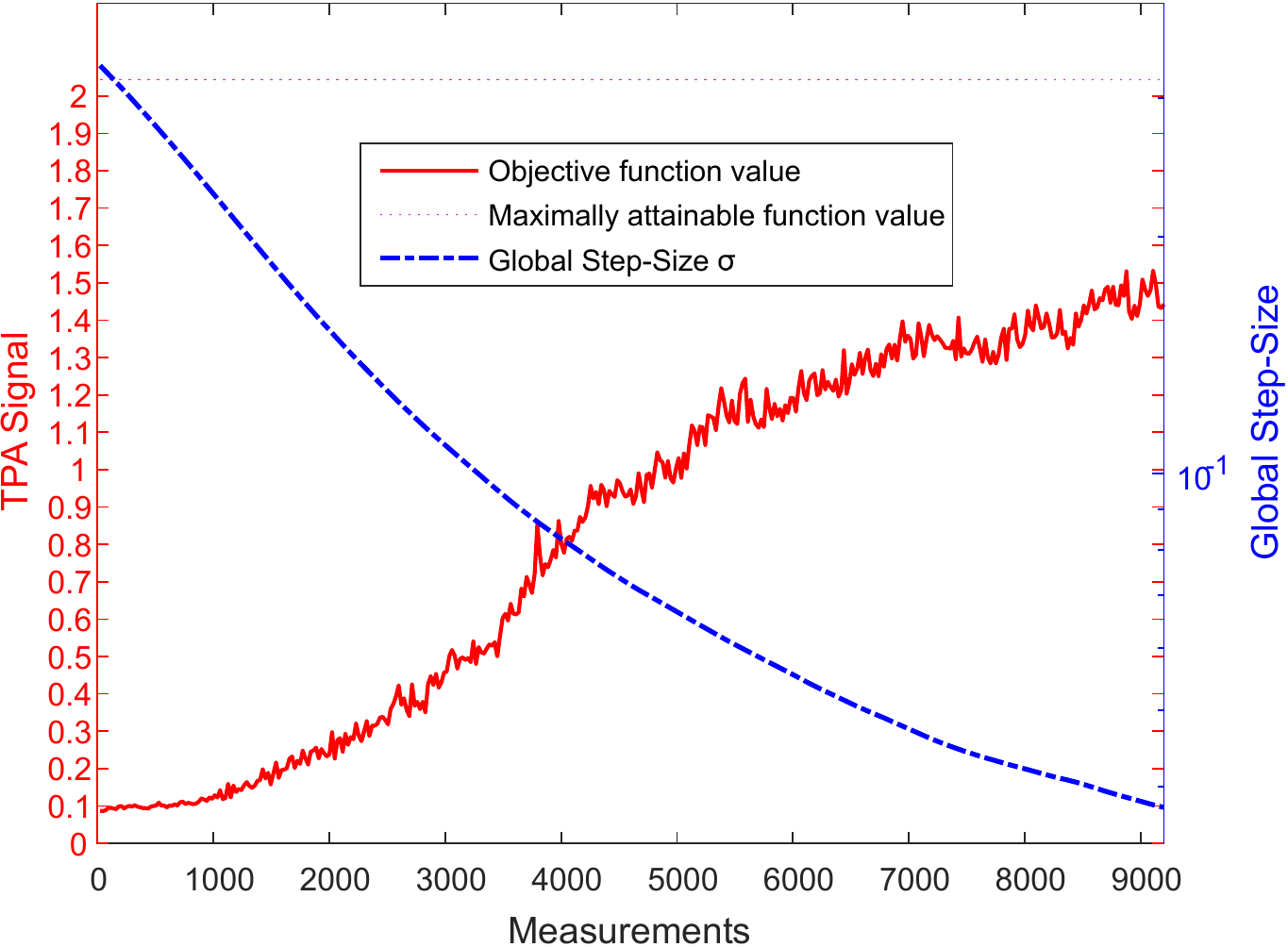}
\includegraphics[width=0.48\columnwidth]{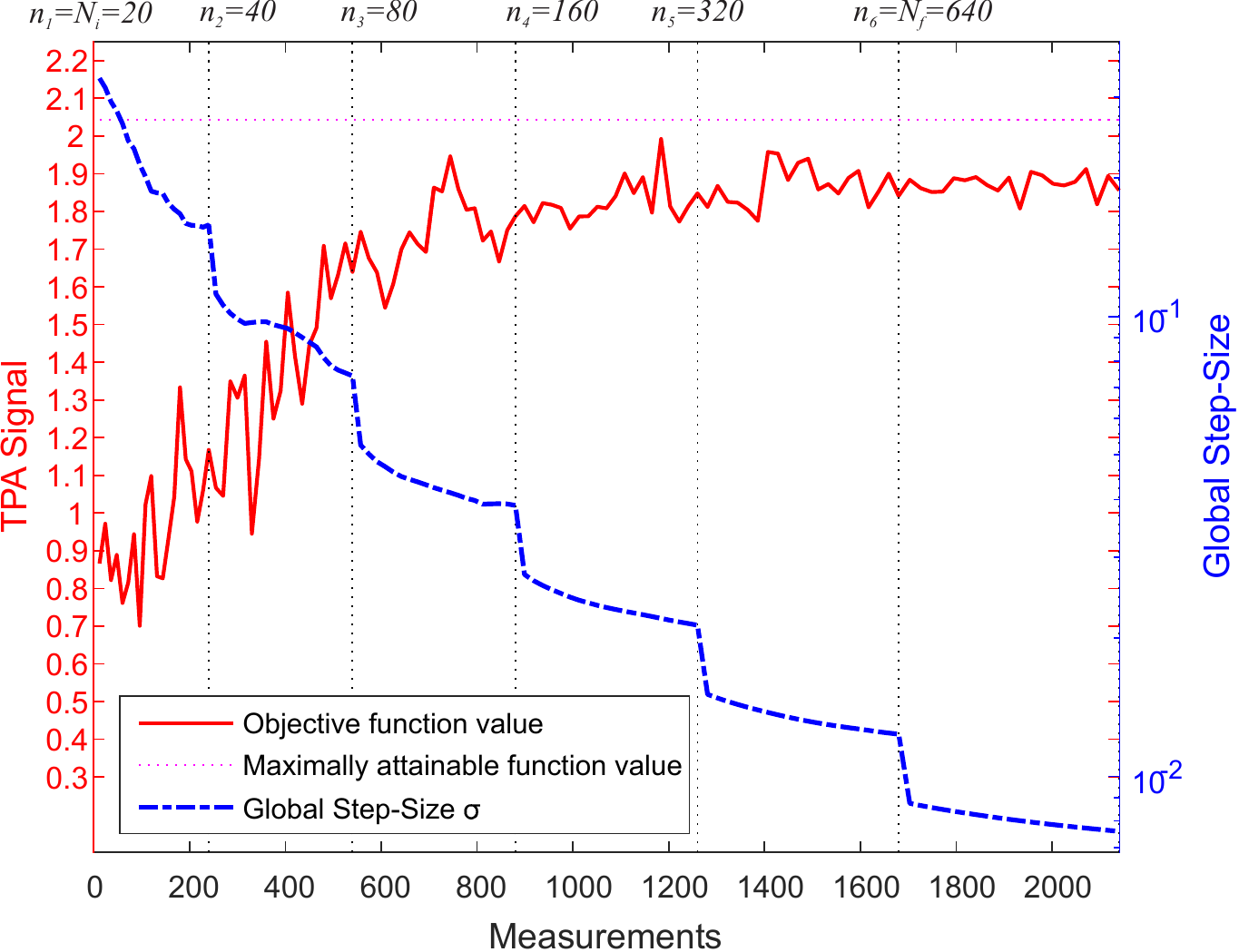}
 \caption{Experimental optimization of the TPA signal in either direct $N_d=640$ controls [LEFT], or in the m-lev approach \textbf{using (tc-0)} [RIGHT].
The objective function values are normalized with respect to the TPA signal from a Transform-Limited pulse (i.e., the flat phase). 
The horizontal dashed line is artificially depicted as a reference representing the maximally attained signal value (2.04). 
This value was independently obtained in a direct search at $N_d=160$, which required at least $5000$ measurements.
 \label{fig:ml_expTPA}}
\end{figure*}

Fig.~\ref{fig:ml_expTPA} depicts typical runs from experimental optimization targeting the maximization of the TPA signal.
The outcome of direct optimization at $N_d=640$ controls, as well as the m-lev approach, are depicted in Fig.~\ref{fig:ml_expTPA}.
In each plot, the red (noisy) curves record the evolution of the objective function values over the entire course of optimization; the curves are the TPA signals normalized with respect to the signal generated by the so-called transform-limited pulse (i.e., recorded from a constant flat phase, $\phi_i=0$ for all $i$ on the shaper). This normalization removes the effect of the laser fluctuations to the fitness over the long period of the experiment, ensuring consistency amongst different runs. The blue curves record the global step-size $\sigma$ during each optimization run. 
The m-lev approach starts from the initial grid size of $N_i=n_1=20$, which doubles in the subsequent cases (i.e., $n_{\ell}=\left\{40,80,160,320\right\}$), and ends up at the finest grid level of $N_f=n_6=640$. 
For each grid level, 20 generations (iterations) are performed, yet the population sizes differ, adhering to Eq.~\ref{eq:popsize}. 
The vertical dashed lines in Fig.~\ref{fig:ml_expTPA}[RIGHT] mark the boundaries of the adjacent grid levels. 
Results from multiple runs suggest that the maximally attainable signal value is in the range of $1.8\sim 2.0$, considering experimental noise of $\approx 10\%$, (i.e., any TPA signal exceeding $1.8$ may be considered as achieving the maximum). 
At the same time, the horizontal dashed lines represent the single measurement of the best attained signal value (2.04; independently obtained in a direct search at $N_d=160$, which required at least $5000$ measurements).
Evidently, the m-lev approach obtained the most efficient run in terms of first reaching the vicinity of the optimum: In this recorded run it required only $\approx 700$ measurements. 
At the same time, the direct search at $N_d=640$ failed to locate the maximum in all the recorded experiments, even after $9000$ measurements. 
Notably, we associate this behavior with the \textit{curse of dimensionality}. 
%The ML approach (Fig.~\ref{fig:ml_expTPA}[RIGHT]) achieved comparable yield values as were obtained in the direct search at $N=160$. % but is far more efficient.

It is also evident from Fig.~\ref{fig:ml_expTPA}[RIGHT] that $n_{\ell}=80$ is the minimally-required grid-size in order to reach the maximum yield. Further increasing the grid-size (e.g., 160, 320, 640) does not exhibit significant yield improvements. 
%Xi suggested to remove: which is primarily governed by the shaper resolution, the laser bandwidth, as well as the nature of the physical observable. 
This additional information is normally not available from the routinely-employed direct runs at a designated dimensionality $N_d$ that is arbitrarily chosen. 
%As expected, the sigma values descent as the optimization proceeds, where the lowest sigma is reached to below 10-2 in the ML approach. 
In summary, the m-lev approach performs best both in terms of yield and speed. 
In fact, its efficiency could be further improved by applying an automated stopping criterion, instead of the utilized (tc-0), which accounts for the inherent noise and uncertainty.

\section{Conclusions} \label{sec:conclusions}
In this study we introduced the notion of m-lev optimization into the realm of self-adaptive ESs \cite{ESchapter2018,EStoRL}. 
We articulated our assumptions, formulated the framework, and devised a set of possible termination criteria for the automated leveling-up.
We proposed instantiations for this framework by means of specific variants, and assessed the \textbf{computational} benefit on simulation-based QC systems as well as a simple experimental system.
We preliminarily investigated the high-dimensional unconstrained quadratic function when artificially placed on a grid, where we also used theoretical results concerning the optimal step-size to analyse the step-size adaptation of the proposed m-lev-1p1 during the course of multi-leveling. Both (tc-1) and (tc-2) proved potent criteria, yet the configuration of (tc-2) exhibited high sensitivity to the search landscape.
We then carried out a systematic testing procedure on a QC simulation-based optimization test-bed, primarily with (tc-1), where the m-lev-sepC performed very well.
Importantly, simulation-based QC systems comprising $2^{14}$ decision variables, which could not be solved in their full-scale formulation within a practical run-time, were efficiently solved by the proposed m-lev ESs.
Interestingly, the proposed m-lev-1p1 outperformed m-lev-sepC on some of the problems.

The utilization of the various upscale operators exhibited different efficacy over the simulated test-cases. 
Nearest-neighbor interpolation (U-1) performed best on two out of the three systems for the two algorithmic instantiations. 
On the third system, all three operators performed equally well.
Generally, given a black-box multi-resolution optimization problem, we expect the fit of the upscale operator to an m-lev-ES to be \textit{problem-dependent}.

Finally, an experimental proof-of-concept of the m-lev-sepC variant on TPA optimization was successfully accomplished in the laboratory, using (tc-0).
The algorithmic benefit was evident as the m-lev approach performed as expected over the 640 control pixels of the pulse-shaper.
The practical significance of applying the m-lev approach will be \textit{problem-dependent} (i.e., application-specific), and the illustrations show the clear prospect of enhancing efficiency.

Primary directions of future work are (i) to investigate flexible \textit{schedules} (e.g., logarithmically reducing the factor of refinement to counteract the \textit{curse of dimensionality}), (ii) to devise and assess additional \textit{upscale operators} in order to improve performance, (iii) to explore additional core solvers (e.g., the limited-memory CMA-ES \cite{LMCMA2014}), (iv) to obtain efficient automated termination that fits zero-assumption black-box settings (e.g., by considering (tc-0) in light of the \textit{multi-armed bandit problem}), and (v) to extend the m-lev notion and operation to 2D grids.

\subsection*{Acknowledgments}
The author H.R.~acknowledges support from the Department of Energy (DE-FG02-02ER15344) and X.X.~acknowledges support from the National Science Foundation (CHE-1763198).

\appendix
\section{The Pixelation Effect}\label{app:pixelation}
\begin{figure*}
\centering
\includegraphics[width=1.0\columnwidth]{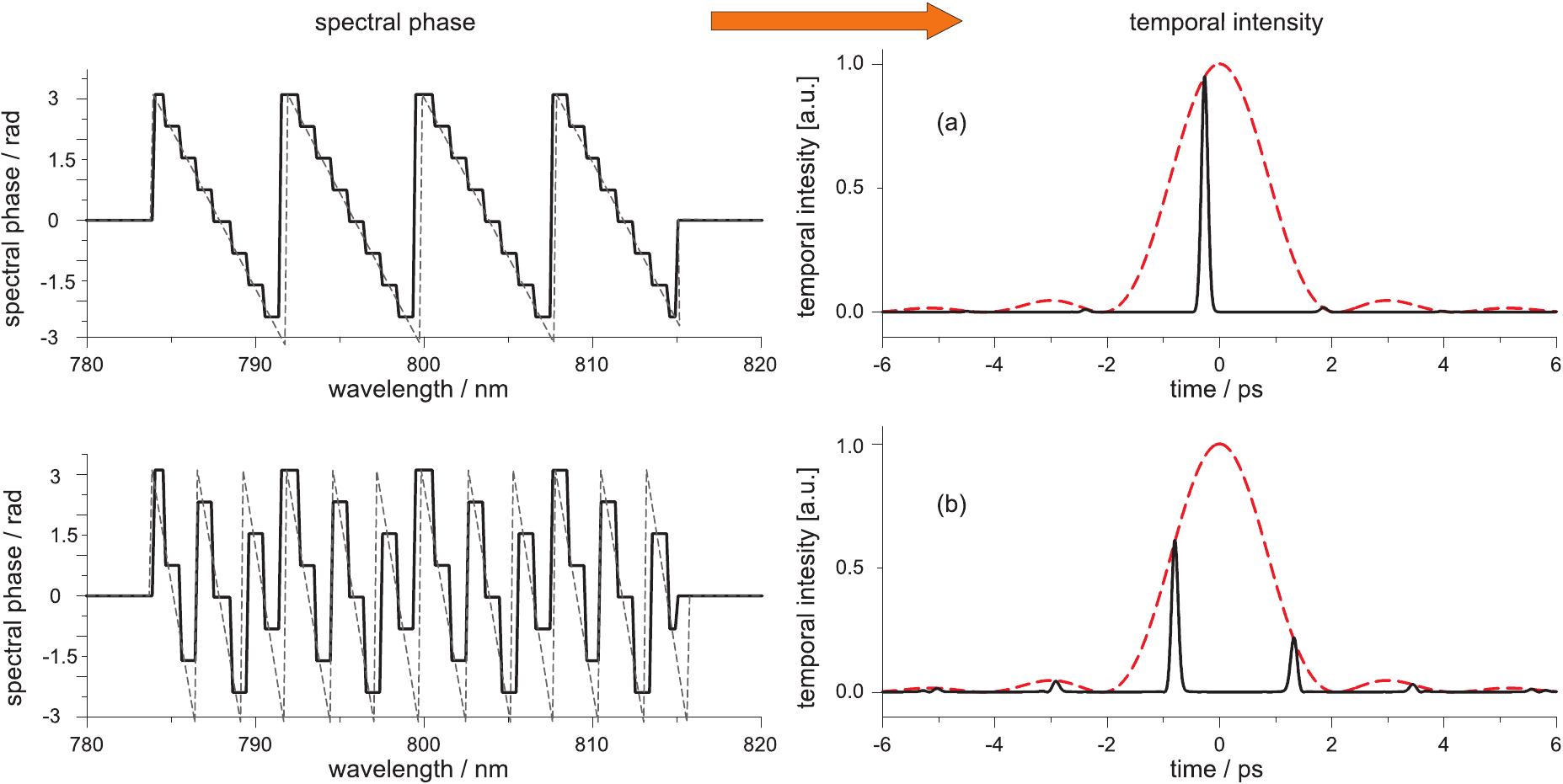}
\caption{Illustration of the pixelation effect occurring in pulse shapers during typical QC experiments; shaping is carried out with 32 pixels in the domain $[-\pi,\pi]$.
Panel (a) presents replica pulses with a moderate linear phase, with the envelope of the replicas in a dashed curve.  Panel (b) shows a steeper phase, with more pronounced replica pulses \cite{MatthiasPHD}.
Figure courtesy of Matthias Roth and Jonathan Roslund.\label{fig:pixelation}}
\end{figure*}

The time modulation of these step-functions is attained by their inverse Fourier transform,
\begin{equation*}
\displaystyle \mathcal{F}^{-1}\left[\textrm{squ}(\nu)\right]\sim
\textrm{sinc}(\tau)
\end{equation*}
where the width of $\textrm{sinc}(\tau)=\frac{\sin(\tau)}{\tau}$ is inversely proportional to the pixel width. 
Explicitly, the resulting temporal electric field in this pixelation can be described by
\begin{equation}
\displaystyle e(t) = \sum_{\jmath} \tilde{e}(t-\jmath \tau)\cdot
\textrm{sinc}\left(\frac{\pi t}{\tau}\right),
\end{equation}
with $\tilde{e}(t)$ as the \emph{desired} electric field, and where $\tau = \frac{1}{\Delta \nu}$ is the inverse frequency spacing per pixel.
Fig.~\ref{fig:pixelation} provides an illustration of the \textit{pixelation} effect \cite{MatthiasPHD}.

\section{Laboratory Setup}\label{app:LabSetup}
A commercial Ti:Sapphire femtosecond laser system consisting of an oscillator and a multi-pass amplifier (KMlab, Dragon) is used to produce pulses with $\sim 25\textrm{fs}$ pulse width centered at $\sim 790\textrm{nm}$. 
The laser pulses are shaped with a dual-mask liquid crystal SLM (CRI, SLM-640), capable of simultaneous phase and amplitude modulation. A small fraction of the shaped laser pulse is spit and focused into a GaP photodiode (Thorlab), which generates only the TPA signal from the $790\textrm{nm}$ incident light. The laser power was adjusted with neutral density filters before entering the photodiode to make sure that the TPA signal does not saturate. 
The SLM contains $N_f=640$ pixels, and the shaper is configured at a resolution of $\sim 0.2\textrm{nm/pixel}$. 
We utilize phase-only modulation in the experiment, with all the amplitudes fixed at 1. 
Adjacent SLM pixels are sequentially tied together in groups of $g_{\ell}~\left(g_{\ell}=32,~16,~8,~4,~2 \right)$ to reduce the 640-dimensional space of spectral phases to the desired grid levels $n_{\ell}~\left(n_{\ell}=20,~40,~80,~160,~320 \right)$, respectively.
The QC objective function is the experimental equivalence of Eq.~\ref{eq:SHG_T}, that is, to experimentally maximize the TPA signal by obtaining the optimal phase mask on the shaper. Ideally, the optimal phase should be a flat phase or a linear phase, but in laboratory practice, it contains higher order nonlinear terms to compensate for or correct those inherent in the laser system.

\begin{figure*}[htb]
\includegraphics[width=1.0\columnwidth]{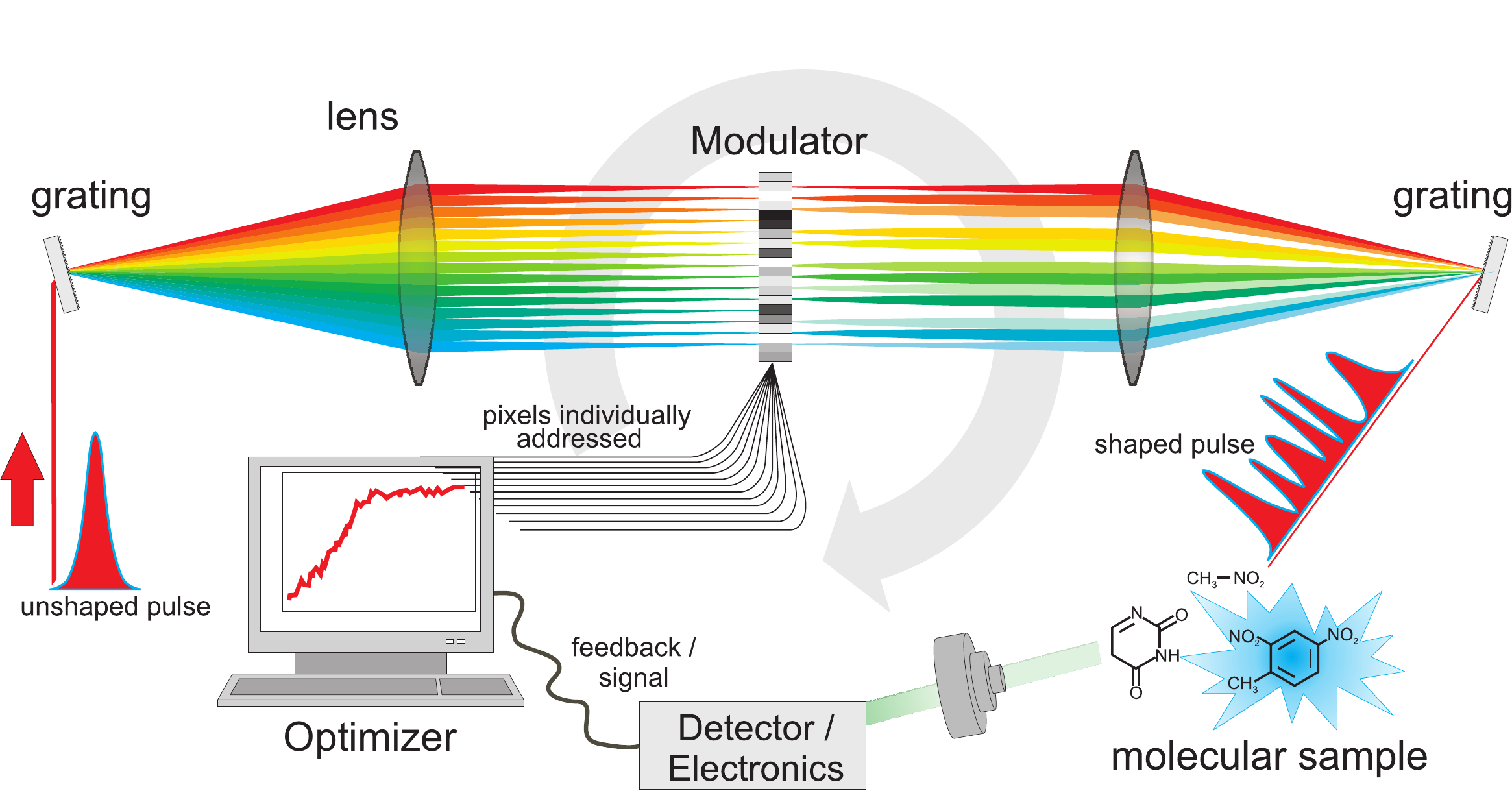}
\caption{The experimental quantum control learning loop. An unshaped laser pulse, approximated well by a Gaussian, is shaped by a pixel-based modulator and applied to a molecular sample. The control variables addressed in the laboratory by the evolution strategy are the individual pixels that determine the pulse shape. The measured signal reflecting the molecular response constitutes the feedback for optimization by the ES. \label{fig:qce}}
\end{figure*}

Fig.~\ref{fig:qce} schematically illustrates an experimental QC learning loop \cite{Gerber07}.

%\bibliographystyle{ieeetr}
%\bibliography{oshir}

\begin{thebibliography}{10}
\providecommand{\url}[1]{{#1}}
\providecommand{\urlprefix}{URL }
\expandafter\ifx\csname urlstyle\endcsname\relax
  \providecommand{\doi}[1]{DOI~\discretionary{}{}{}#1}\else
  \providecommand{\doi}{DOI~\discretionary{}{}{}\begingroup
  \urlstyle{rm}\Url}\fi

\bibitem{ArnoldNoise}
Arnold, D.V., Beyer, H.G.: {Local Performance of the $(\mu/\mu_I, \lambda)$-ES
  in a Noisy Environment}.
\newblock In: W.~Martin, W.~Spears (eds.) {Foundations of Genetic Algorithms,
  6}, pp. 127--141. Morgan Kaufmann, San Francisco, CA (2001)

\bibitem{Arnold2006_HOES}
Arnold, D.V., MacLeod, A.: Hierarchically organised evolution strategies on the
  parabolic ridge.
\newblock In: Proceedings of the 8th Annual Conference on Genetic and
  Evolutionary Computation, GECCO '06, pp. 437--444. ACM, New York, NY, USA
  (2006).
\newblock \doi{10.1145/1143997.1144080}.
\newblock \urlprefix\url{http://doi.acm.org/10.1145/1143997.1144080}

\bibitem{MultigridGAs}
Asbury, S., Holloway, J.P.: Multi-grid genetic algorithms for space shield
  design.
\newblock In: Proceedings of the International Conference on Mathematics,
  Computational Methods and Reactor Physics (2009)

\bibitem{Baeck-book}
B{\"a}ck, T.: {E}volutionary {A}lgorithms in {T}heory and {P}ractice.
\newblock Oxford University Press, New York, NY, USA (1996)

\bibitem{Baeck2013contemporary}
B{\"a}ck, T., Foussette, C., Krause, P.: Contemporary Evolution Strategies.
\newblock Natural Computing Series. Springer-Verlag Berlin Heidelberg (2013)

\bibitem{barth2001multiscale}
Barth, T., Chan, T., Haimes, R.: Multiscale and Multiresolution Methods: Theory
  and Applications.
\newblock Lecture Notes in Computational Science and Engineering.
  Springer-Verlag Berlin Heidelberg (2002)

\bibitem{Bellman1961}
Bellman, R.E.: Adaptive Control Processes: A Guided Tour.
\newblock Princeton University Press (1961)

\bibitem{Beyer93}
Beyer, H.G.: {{T}oward a {T}heory of {E}volution {S}trategies: Some
  {A}symptotical {R}esults from the $(1\stackrel{+}{,}\lambda)$-{T}heory}.
\newblock Evolutionary Computation \textbf{1}(2), 165--188 (1993)

\bibitem{Beyer97}
Beyer, H.G.: An {A}lternative {E}xplanation for the {M}anner in which {G}enetic
  {A}lgorithms {O}perate.
\newblock BioSystems \textbf{41}(1), 1--15 (1997)

\bibitem{Beyer}
Beyer, H.G.: {T}he {T}heory of {E}volution {S}trategies.
\newblock Springer, Heidelberg (2001)

\bibitem{Boyd}
Boyd, S., Vandenberghe, L.: Convex Optimization.
\newblock Cambridge University Press, New York (2004)

\bibitem{MultigridSolvers}
Brandt, A., Ron, D.: Multigrid solvers and multilevel optimization strategies.
\newblock In: J.~Cong, J.R. Shinnerl (eds.) Multilevel Optimization in VLSICAD,
  \emph{Combinatorial Optimization}, vol.~14, pp. 1--69. Springer US (2003).
\newblock \doi{10.1007/978-1-4757-3748-6_1}.
\newblock \urlprefix\url{http://dx.doi.org/10.1007/978-1-4757-3748-6_1}

\bibitem{Branke}
Branke, J.: {Evolutionary Optimization in Dynamic Environments}.
\newblock Kluwer Academic Publishers, Norwell, MA, USA (2001)

\bibitem{MultigridTutorial}
Briggs, W.L., Henson, V.E., McCormick, S.F.: {A Multigrid Tutorial}, 2nd edn.
\newblock Society for Industrial and Applied Mathematics, Philadelphia, PA, USA
  (2000)

\bibitem{DantzigWolfe}
Dantzig, G.B., Wolfe, P.: Decomposition principle for linear programs.
\newblock Operations Research \textbf{8}(1), 101--111 (1960).
\newblock \doi{10.1287/opre.8.1.101}.
\newblock \urlprefix\url{http://dx.doi.org/10.1287/opre.8.1.101}

\bibitem{ESchapter2018}
Emmerich, M., Shir, O.M., Wang, H.: Handbook of Heuristics, chap. Evolution
  Strategies, pp. 1--31.
\newblock Springer International Publishing, Cham (2018).
\newblock \doi{10.1007/978-3-319-07153-4_13-1}.
\newblock \urlprefix\url{https://doi.org/10.1007/978-3-319-07153-4_13-1}

\bibitem{Giannakoglou2010}
Giannakoglou, K.C., Kampolis, I.C.: Computational Intelligence in Expensive
  Optimization Problems, chap. Multilevel Optimization Algorithms Based on
  Metamodel- and Fitness Inheritance-Assisted Evolutionary Algorithms, pp.
  61--84.
\newblock Springer Berlin Heidelberg, Berlin, Heidelberg (2010).
\newblock \doi{10.1007/978-3-642-10701-6_3}.
\newblock \urlprefix\url{http://dx.doi.org/10.1007/978-3-642-10701-6_3}

\bibitem{hackbusch1985multi}
Hackbusch, W.: Multi-Grid Methods and Applications.
\newblock Springer series in computational mathematics. Springer-Verlag (1985)

\bibitem{hansen2015}
Hansen, N., Arnold, D.V., Auger, A.: Springer Handbook of Computational
  Intelligence, chap. Evolution Strategies, pp. 871--898.
\newblock Springer Berlin Heidelberg, Berlin, Heidelberg (2015).
\newblock \doi{10.1007/978-3-662-43505-2_44}.
\newblock \urlprefix\url{http://dx.doi.org/10.1007/978-3-662-43505-2_44}

\bibitem{Harvey92speciesadaptation}
Harvey, I.: Species adaptation genetic algorithms: A basis for a continuing
  saga.
\newblock In: Proceedings of the First European Conference on Artificial Life,
  pp. 346--354. MIT Press/Bradford Books (1992)

\bibitem{MultigridEAs}
He, J., Kang, L.: A mixed strategy of combining evolutionary algorithms with
  multigrid methods.
\newblock International Journal of Computer Mathematics \textbf{86}(5),
  837--849 (2009).
\newblock \doi{10.1080/00207160701713581}

\bibitem{Koza1990GP}
Koza, J.R.: Genetic programming: A paradigm for genetically breeding
  populations of computer programs to solve problems.
\newblock Tech. rep., Stanford University, Stanford, CA, USA (1990)

\bibitem{LaForgeMO}
Laforge, F., Roslund, J., Shir, O.M., Rabitz, H.: Multiobjective adaptive
  feedback control of two-photon absorption coupled with propagation through a
  dispersive medium.
\newblock Phys. Rev. A \textbf{84}, 013401 (2011).
\newblock \doi{10.1103/PhysRevA.84.013401}

\bibitem{smac-2017}
Lindauer, M., Eggensperger, K., Feurer, M., Falkner, S., Biedenkapp, A.,
  Hutter, F.: Smac v3: Algorithm configuration in python.
\newblock \url{https://github.com/automl/SMAC3} (2017)

\bibitem{EAsMultifidelity2016}
Liu, B., Koziel, S., Zhang, Q.: A multi-fidelity surrogate-model-assisted
  evolutionary algorithm for computationally expensive optimization problems.
\newblock Journal of Computational Science \textbf{12}, 28 -- 37 (2016).
\newblock \doi{https://doi.org/10.1016/j.jocs.2015.11.004}

\bibitem{IRACE}
L{\'o}pez-Ib{\'a}{\~n}ez, M., Dubois-Lacoste, J., {P{\'e}rez C{\'a}ceres}, L.,
  St\"{u}tzle, T., Birattari, M.: The {{irace}} package: Iterated racing for
  automatic algorithm configuration.
\newblock Operations Research Perspectives \textbf{3}, 43--58 (2016).
\newblock \doi{10.1016/j.orp.2016.09.002}

\bibitem{LMCMA2014}
Loshchilov, I.: A computationally efficient limited memory cma-es for large
  scale optimization.
\newblock In: Proceedings of the 2014 Annual Conference on Genetic and
  Evolutionary Computation, GECCO '14, p. 397–404. Association for Computing
  Machinery, New York, NY, USA (2014).
\newblock \doi{10.1145/2576768.2598294}

\bibitem{Multifidelity2012}
March, A., Willcox, K.: Provably convergent multifidelity optimization
  algorithm not requiring high-fidelity derivatives.
\newblock AIAA Journal \textbf{50}(5), 1079--1089 (2012).
\newblock \doi{10.2514/1.J051125}

\bibitem{Hall2016}
{Mathew}, R., {Hall}, K.C.: {Tailoring the ultrafast control of quantum dot
  excitons using optical pulse shaping}.
\newblock Physica Status Solidi C Current Topics \textbf{13}, 67--72 (2016).
\newblock \doi{10.1002/pssc.201510152}

\bibitem{Gerber07}
Nuernberger, P., Vogt, G., Brixner, T., Gerber, G.: {Femtosecond Quantum
  Control of Molecular Dynamics in the Condensed Phase}.
\newblock Phys Chem Chem Phys. \textbf{9}(20), 2470--2497 (2007)

\bibitem{powell2013optimal}
Powell, W., Ryzhov, I.: Optimal Learning.
\newblock Wiley Series in Probability and Statistics. Wiley (2013)

\bibitem{NumericalRecipes}
Press, W., Teukolsky, S., Vetterling, W., Flannery, B.: {Numerical Recipes in
  C}, 2nd edn.
\newblock Cambridge University Press, Cambridge, UK (1992)

\bibitem{Hersch00}
Rabitz, H., de~Vivie-Riedle, R., Motzkus, M., Kompa, K.: {Whither the Future of
  Controlling Quantum Phenomena?}
\newblock Science \textbf{288}, 824--828 (2000)

\bibitem{Rechenberg}
Rechenberg, I.: {Evolutionsstrategies: Optimierung technischer Systeme nach
  Prinzipien der biologischen Evolution}.
\newblock Frommann-Holzboog Verlag, Stuttgart, Germany (1973)

\bibitem{HansenDR2PPSN08}
Ros, R., Hansen, N.: {A} {S}imple {M}odification in {CMA}-{ES} {A}chieving
  {L}inear {T}ime and {S}pace {C}omplexity.
\newblock In: Parallel Problem Solving from Nature - PPSN X, \emph{Lecture
  Notes in Computer Science}, vol. 5199, pp. 296--305. Springer (2008)

\bibitem{Vrakking}
Rosca-Pruna, F., Vrakking, M.J.: {Revival Structures in Picosecond
  Laser-Induced Alignment of I2 Molecules}.
\newblock Journal of Chemical Physics \textbf{116}(15), 6579--6588 (2002)

\bibitem{LabES}
Roslund, J., Shir, O.M., B\"ack, T., Rabitz, H.: {Accelerated Optimization and
  Automated Discovery with Covariance Matrix Adaptation for Experimental
  Quantum Control}.
\newblock Physical Review A (Atomic, Molecular, and Optical Physics)
  \textbf{80}(4), 043415 (2009).
\newblock \doi{10.1103/PhysRevA.80.043415}

\bibitem{MatthiasPHD}
Roth, M.: {Optimal Dynamic Discrimination in the Laboratory}.
\newblock Ph.D. thesis, Princeton University (2007)

\bibitem{EStoRL}
Salimans, T., Ho, J., Chen, X., Sutskever, I.: Evolution strategies as a
  scalable alternative to reinforcement learning.
\newblock CoRR \textbf{abs/1703.03864} (2017).
\newblock \urlprefix\url{http://arxiv.org/abs/1703.03864}

\bibitem{GECCO07_SHG}
Shir, O.M., B{\"a}ck, T.: The {S}econd {H}armonic {G}eneration {C}ase {S}tudy
  as a {G}ateway for {ES} to {Q}uantum {C}ontrol {P}roblems.
\newblock In: Proceedings of the Genetic and Evolutionary Computation
  Conference, GECCO-2007, pp. 713--721. ACM Press, New York, NY, USA (2007)

\bibitem{SeqExpEA_TutorialGECCO2018}
Shir, O.M., B\"{a}ck, T.: Sequential experimentation by evolutionary
  algorithms.
\newblock In: Proceedings of the Genetic and Evolutionary Computation
  Conference Companion, GECCO '18, pp. 956--976. ACM, New York, NY, USA (2018).
\newblock \doi{10.1145/3205651.3207885}.
\newblock \urlprefix\url{http://doi.acm.org/10.1145/3205651.3207885}

\bibitem{Shir-JPhysB}
Shir, O.M., Beltrani, V., B{\"a}ck, T., Rabitz, H., Vrakking, M.J.: {On the
  Diversity of Multiple Optimal Controls for Quantum Systems}.
\newblock Journal of Physics B: Atomic, Molecular and Optical Physics
  \textbf{41}(7), 074021 (2008).
\newblock \doi{10.1088/0953-4075/41/7/074021}

\bibitem{QCE_GECCO08}
Shir, O.M., Roslund, J., B{\"a}ck, T., Rabitz, H.: {Performance Analysis of
  Derandomized Evolution Strategies in Quantum Control Experiments}.
\newblock In: Proceedings of the 10th Genetic and Evolutionary Computation
  Conference, GECCO-2008, pp. 519--526. ACM Press, New York, NY, USA (2008)

\bibitem{Shir-MOQC}
Shir, O.M., Roslund, J., Leghtas, Z., Rabitz, H.: Quantum control experiments
  as a testbed for evolutionary multi-objective algorithms.
\newblock Genetic Programming and Evolvable Machines \textbf{13}, 445--491
  (2012).
\newblock \urlprefix\url{http://dx.doi.org/10.1007/s10710-012-9164-7}

\bibitem{Tilahun2012}
Tilahun, S.L., Kassa, S.M., Ong, H.C.: PRICAI 2012: Trends in Artificial
  Intelligence: 12th Pacific Rim International Conference on Artificial
  Intelligence, Kuching, Malaysia, September 3-7, 2012. Proceedings, chap. A
  New Algorithm for Multilevel Optimization Problems Using Evolutionary
  Strategy, Inspired by Natural Adaptation, pp. 577--588.
\newblock Springer Berlin Heidelberg, Berlin, Heidelberg (2012).
\newblock \doi{10.1007/978-3-642-32695-0_51}.
\newblock \urlprefix\url{http://dx.doi.org/10.1007/978-3-642-32695-0_51}

\bibitem{Hersch93}
Warren, W.S., Rabitz, H., Dahleh, M.: {Coherent Control of Quantum Dynamics:
  The Dream Is Alive}.
\newblock Science \textbf{259}, 1581--1589 (1993)

\bibitem{MultigridAIS}
Watanabe, K., Campelo, F., Igarashi, H.: Topology optimization based on immune
  algorithm and multi grid method.
\newblock IEEE Transactions on Magnetics \textbf{43}(4) (2007)

\bibitem{Weiner00}
Weiner, A.M.: Femtosecond pulse shaping using spatial light modulators.
\newblock Review of Scientific Instruments \textbf{71}(5), 1929--1960 (2000).
\newblock \doi{10.1063/1.1150614}

\end{thebibliography}

\end{document}